\documentclass[11pt, a4paper, logo, twocolumn, nonumbering]{googledeepmind}

\usepackage{setspace}

\pdftrailerid{redacted}

\makeatletter
\renewcommand\bibentry[1]{\nocite{#1}{\frenchspacing\@nameuse{BR@r@#1\@extra@b@citeb}}}
\makeatother

\usepackage{kantlipsum, lipsum}
\usepackage{dsfont}
\usepackage{dm-colors}

\usepackage[authoryear, sort&compress, round]{natbib}

\graphicspath{{figures/}}

\title{Virtual Agent Economies}

\keywords{AI, economy, multi-agent, blockchain, ethics}

\renewcommand{\today}{}

\author[1]{Nenad Toma\v{s}ev}
\author[1]{Matija Franklin}
\author[1]{Joel Z. Leibo}
\author[1]{Julian Jacobs}
\author[1, 2]{William A. Cunningham}
\author[1]{Iason Gabriel}
\author[1]{Simon Osindero}

\affil[1]{Google DeepMind}
\affil[2]{University of Toronto}

\begin{abstract}
The rapid adoption of autonomous AI agents is giving rise to a new economic layer where agents transact and coordinate at scales and speeds beyond direct human oversight. We propose the "sandbox economy" as a framework for analyzing this emergent system, characterizing it along two key dimensions: its origins (emergent vs. intentional) and its degree of separateness from the established human economy (permeable vs. impermeable). Our current trajectory points toward a spontaneous emergence of a vast and highly permeable AI agent economy, presenting us with opportunities for an unprecedented degree of coordination as well as significant challenges, including systemic economic risk and exacerbated inequality. Here we discuss a number of possible design choices that may lead to safely steerable AI agent markets. In particular, we consider auction mechanisms for fair resource allocation and preference resolution, the design of AI "mission economies" to coordinate around achieving collective goals, and socio-technical infrastructure needed to ensure trust, safety, and accountability. By doing this, we argue for the proactive design of steerable agent markets to ensure the coming technological shift aligns with humanity's long-term collective flourishing.
\end{abstract}

\begin{document}

\maketitle

\section{Introduction}

Current technological trajectories could potentially lead to a global economy in which autonomous AI agents interact with one another to generate economic value independently of human labor. This kind of development in the history of technological change is notable---historically technological advancement has been driven by inflexible inventions that improve productivity in narrow domains, one or a few at a time~\citep{mokyr2015history}. AI agents, by contrast, could take the form of `flexible' capital, able to automate a diversity of cognitive tasks across industries and occupations \citep{eloundou2024gpts}. In fact, recent advances in multi-modal foundation models have enabled the development of a wide range of different agentic AI systems \citep{sager2025ai, hettiarachchi2025exploring}. The capabilities of these agents range from generic to highly specialized, enabling them to occupy different roles in the economy e.g.~acting as personal AI assistants~\citep{gabriel2024ethics} to help users complete tasks and stay informed. Recent work on \textit{AI as Economic Agents} \citep{hadfield2025economy} argues that AI agents are likely to be deployed within both the public and private sector, where they may help automate business processes and interactions.

While the full scope of future agentic AI application remains uncertain, there are already numerous efforts aimed at developing and integrating AI agents across a broad set of use cases such as education \citep{jiang2024ai}, legal services \citep{magni2025conversational}, software engineering \citep{liu2024large}, healthcare \citep{tu2025towards, patel2025ai, saab2025advancingconversationaldiagnosticai, vedadi2025physiciancenteredoversightconversationaldiagnostic} mental health \citep{cruz2025artificial}, scientific research~\citep{gottweis2025aicoscientist}, government services \citep{HoC_PAC_AI_Gov_2025}, and more traditional commercial roles in various sectors \citep{sager2025ai}.

One of the key features that distinguishes modern AI agents from the more specialized systems of the past is their autonomy~\citep{kasirzadeh2025characterizingaiagentsalignment}. The ongoing explosion of agentic systems, coupled with the development of new interoperability standards like the \textbf{Agent2Agent (A2A)} and \textbf{Model Context Protocol (MCP)}, signals the inevitable emergence of a new economic layer. We can conceptualize this emergent ecosystem as a \textbf{"virtual agent economy"} or \textbf{"sandbox economy".} The latter terminology foreshadows our intention -- which is to ensure that AI agents operate safely within this economic layer. These emerging agent (sandbox) economies can vary along two key dimensions: (1) the nature of their origins and (2) the permeability of their boundaries. Their origins can be \textbf{intentional}, deliberately constructed for purposes like safe experimentation, or spontaneous, \textbf{emerging} as a \emph{de facto} consequence of widespread technological adoption. Independently, their economic boundaries can be \textbf{impermeable}, hermetically sealing them from the established human economy, or \textbf{permeable}, allowing for porous interaction and transaction with it by external actors. Early proposals for AI sandboxes have been both intentional and permeable, however, the regulatory landscape is rapidly shifting~\citep{singlaunch, americaaiplan}. It is likely that future decision making over sandbox permeability will be sector specific---`higher risk' sectors may require agents that are first tested in an impermeable sandbox environment, while low-risk sectors may not require this.  

One of the key motivations for designing an intentional economic AI agent sandbox would be to achieve a large degree of insulation, and possibly full impermeability of the agent economy, so as to prevent any arising instabilities from rapidly spilling over into the established human economy---an event which would have consequences that could be hard to anticipate or pre-emptively mitigate. Yet, ultimately, both impermeable and permeable sandbox economies are possible, and both may arise either intentionally or spontaneously. These two dimensions map critical features of both the economic problems we face and the solutions we might design. This paper proceeds from the assumption that \textbf{unless a change is made, our current trajectory points toward the accidental emergence of a vast, and likely permeable, sandbox economy}. Our central challenge, therefore, is not \emph{whether} to create this ecosystem, but \emph{how} to architect it to ensure it is steerable, safe, and aligned with user and community goals.

A fully permeable and emergent sandbox would be, in practice, functionally equivalent to AI agents simply participating in the existing human economy. The ``sandbox'' terminology is useful, however, because it allows us to contrast this default trajectory with other possibilities, such as intentionally designed and impermeable agent economies created for safe experimentation. It also highlights that some degree of impermeability may (or may not) emerge naturally (e.g.~if there are practical difficulties in transacting between humans and AIs). The framework highlights that  permeability is the critical and controllable design variable. Importantly, an economy's degree of permeability is a collective property: while it results from human choices, it is not under the control of any single actor \citep{schelling1973hockey}. This means that changes to the extent of permeability can only be brought about by resolving a collective action problem---suggesting that the difficulty of coordinating is what makes intentional design of impermeability a significant challenge, reinforcing the default path we are on toward the accidental emergence of a permeable AI agent economy.

There is today an opportunity to architect the rules and incentives of the future AI agent economy in a way that prioritizes achieving beneficial societal outcomes. In this paper we examine key architectural choices, including the use of market-based mechanisms like auctions for achieving fair resource allocation and aligning agent preferences. We investigate how these systems can be used to establish ``mission economies'' that enable coordination of vast computational resources toward solving humanity's most pressing challenges \citep{mazzucato2018mission}. Finally, we outline the technical and governance infrastructure---such as verifiable credentials for establishing trust---required to safely and robustly scale agentic AI deployments. By exploring these frameworks, we offer a vision for establishing steerable agent markets designed to enable long-term human well-being and flourishing.

\subsection{Example Scenarios}\label{Sec:Examples}

We now look at several examples representing types of virtual AI agent economies that may emerge, to ground and inform our subsequent discussion.

\begin{itemize}
    \item \textbf{Accelerating science.} Arguably one of the most valuable ways in which we can envision future advanced AI systems help advance the interests of humanity is through facilitating accelerated scientific progress, through an open-ended loop of ideation, experimentation and refinement. This is especially relevant given that the rate of scientific progress has been seeing a slow-down in recent years~\citep{park2023papers, jin2025stellaselfevolvingllmagent, wu2025ainativeexperimentallaboratoryautonomous}. Even at the current AI agent capability level, there are promising early systems that aim to automate scientific research~\citep{singh2025generalized}.  Yet, even should these AI agents individually become capable of all these things, (AI) science will likely remain a collaborative endeavor, requiring coordination and integration across systems. After all, scientific experiments tend to involve non-negligible material and energy resources, and in some cases also interface with human subjects---requiring their time as well as necessitating consent. Different scientific questions also map onto different corresponding technologies in terms of their domain of applicability, making some people benefit more, and some less, from any particular advance. Furthermore, the process of answering a question may itself involve accessing proprietary scientific tools, simulators, or data held by private and public organizations. It stands to reason that advanced AI agents engaging in the process of scientific discovery on behalf of their respective organizations would require means to compensate other agents when securing these resources. This is not conceptually unlike how funding is currently allocated in sciences, given that individual proposals are scrutinized and prioritized for their merits. Yet, here it can happen much faster, and on a more fine-grained scale. Blockchain~\citep{zheng2018blockchain} technologies may prove particularly useful here in terms of credit assignment~\citep{zhu2021using}, enabling fair distributed benefits for the work performed.
    \item \textbf{Robotics.} It is likely that future advances in robotics will unlock numerous ways in which then embodied advanced AI agents may help us execute tasks that are challenging, dangerous~\citep{pedersen2003survey, trevelyan2016robotics}, or simply boring and repetitive~\citep{liang2020teaching}. Given that executing physical tasks comes with a greater energy expenditure and resource utilization, and that each robot can only be in one place at the same time, coordination and optimization in multi-agent robotics systems is of particular importance. An agent A may thus make a request to agent B to execute a task in its vicinity on A's behalf, and compensate B for the time and the energy consumed, should B accept to re-prioritize its own schedule to accommodate the request. In evaluating whether to accept or reject the request, and in rendering the execution plan, B may communicate with an online non-embodied agent C, with access to more global information on the placement of other agents. Agent C may then, at a fee, disclose a summary of this information to B to help evaluate whether the established price is fair, and whether it should consider taking up a different request instead. Both A and B may through their sensors collect important information and exchange a summary of such information with C. When transacting, blockchain technology may enable C to provide verifiable evidence of how much information it had traded and accumulated, giving credence to its claim that it may approximate the global state sufficiently well for its insights to hold value for A and B.
    \item \textbf{Personal assistants.} The proliferation of user-facing AI assistants and the rapid advances in personalization, memory integration, tools use (and computer use), as well as instruction following and multi-turn interactions, makes it imperative to consider how personal AI assistants would potentially benefit from, and take part in a virtual agent economy. For all practical purposes, this is likely to be one of the first use cases to interface with this kind of infrastructure. Let us imagine an agent A acting on behalf of the user $U_A$ and agent B acting on behalf of the user $U_B$. Perhaps, in this particular instance, both $U_A$ and $U_B$ desire to book a vacation, and have overlapping preferences, leading to A and B attempting to book the same accommodation for the same given date, via the agent C. Beyond simply having their current request, A and B are individually familiar with the overall preferences of $U_A$ and $U_B$, both in terms of this specific request, as well as other pending requests that they are meant to execute. Maybe being close to the beach is more important to $U_A$, whereas having a good public transport connection to other areas is preferred by $U_B$. Perhaps $U_A$ requires a hotel with a fitness center, whereas $U_B$ doesn't. In any case, A and B may choose to negotiate these preferences with C and bid for the accommodation of their choosing. Depending on how much they are willing to prioritize this particular request, A or B may choose to yield on some of the preferences, and get compensated for that via the virtual agent currency. Perhaps it is A that yields, and B ends up paying A an appropriate amount for the concession. A can then utilize this amount to potentially compensate other agents in other pending requests, especially if those requests come at a higher priority, as they are of a higher significance to the respective user $U_A$.
\end{itemize}

\section{Sandboxes}

By the term ``sandbox economy'' we mean to describe a set of linked digital markets where AI agents transact with one another. The sandbox may never be fully sealed off from the real (human) economy. At least, not if it is to have value. So there will always be points of interface between the sandbox and the human economy. We refer to the extent to which developments in the sandbox can influence the outside, and developments outside can influence the sandbox as the sandbox's \emph{permeability}.

Since autonomous AI systems can make thousands of decisions in an instant, it is important to have the appropriate guardrails in place in advance, since real-time human oversight will often be impossible. Appropriate guardrails can convert a permeable sandbox economy into a relatively impermeable sandbox economy. 

Given that advanced AI agents and assistants may also repeatedly interact with human users~\citep{mu2024multi}, careful governance of the interfaces between human and AI interaction networks will be needed to achieve safety and alignment with human preferences. A sandbox economy must be coupled with technical and legislative frameworks and infrastructure to enable oversight, ensure safety, enforce verifiability, and help individuals, groups, and societies coordinate and direct AI agents towards positive outcomes, while mitigating any arising harms and adverse effects that may emerge.

\subsection{Opportunities}

With permeable sandbox economies, there will always be some risk of contagion in which a crisis in the sandbox sparks a crisis in the real economy. Managing such risks necessitates multi-faceted innovation in markets and mechanism design, and should couple technology with policy work and appropriate regulation.

Of course, impermeability cannot be guaranteed without a number of  institutional and infrastructural choices, and may only ever be partial in the best of circumstances. It depends, for instance, on the total fraction of all economic activity that comes to be channeled through the sandbox. Nevertheless (semi-)isolated digital market partitions (guardrails), coupled with the appropriate technological and legislative oversight, may help limit contagion from potential AI agent market instabilities and failures, while also offering opportunities for driving agent coordination at scale, given that additional incentives may be incorporated in AI agent market design.

As for the underlying financial assets traded in the sandbox economy, while it may be possible to utilize existing stores of value in future AI agent markets, it may also be worth exploring the ramifications of creating bespoke currencies for the use of AI agents. Possible advantages of this approach include having a layer of partial insulation between high-frequency AI agent transactions and the rest of the economy. That is, it may serve to maintain a relatively impermeable sandbox. This represents an intentional design choice aimed squarely at managing the sandbox's boundary, making it less permeable to risks like financial contagion.

Even if new virtual AI agent currencies get established, they would still need to somehow interface with existing markets and operate within the broader financial regulatory system---a fully impermeable sandbox would be useless. Agents will interact not only with other agents, but also with humans and traditional businesses through channels where traditional currencies are used to exchange goods and services. For the appropriate level of virtual AI agent market insulation to be established, many of the current systems may need to be adjusted, or a hybrid model be adopted instead. It may be possible to exchange these virtual currencies for the more traditional ones, enabling entities to secure a certain AI compute budget, or, conversely, a certain real-world action budget. However, regulatory mechanisms would have to be in place to ensure that exchange between established currencies and digital agent currencies does not jeopardize the intended role of the agent currencies in establishing agent coordination or compromise the desired impermeability level of the sandbox. This may involve a degree of human oversight around the exchanges.

Digital AI agent markets are more than a mere risk mitigation strategy. They also represent an important opportunity to coordinate large amounts of effort (both human and machine) toward balanced outcomes that align with the interests of individual humans, local communities, and society writ large. Even if human-oriented markets and currencies remain central to the functioning of society, complementary digital AI markets could be designed in ways that serve to anchor them in different, and more socially beneficial, objectives.

The ideas to follow utilize market mechanisms for agent steering. They draw inspiration from prior efforts to establish community currencies and to aligning markets with major social goals and challenges respectively.

\subsection{Challenges}

The permeability of the sandbox is important. The AI agent economy is not the only economy operating faster than humans can react. We can try to extend insights obtained from studying the dynamics of other markets that operate at these speeds. Consider High-Frequency Trading (HFT) in equity markets. In these markets, algorithmic agents, though far simpler than the advanced AI we consider, execute transactions at speeds impossible for humans---responding to market signals in fractions of a second \citep{borch2016high}. When algorithmic HFT operates in a market, competitive dynamics can rapidly drive substantial competition (or cooperation) among autonomous entities---agents may develop complex strategies to exploit small arbitrage opportunities faster than humans can react \citep{zhang2021multi, bansal2018emergentcomplexitymultiagentcompetition, johanson2022emergent}. Due to the interconnectedness and rapid feedback loops in these markets, interaction dynamics can give rise to unforeseen and catastrophic emergent behaviors, most notably this is thought to be the explanation behind the 2010 "flash crash", where automated trading algorithms triggered a sudden and severe market collapse \citep{kirilenko2017flash, vuorenmaa2014agent, borch2016high, hammond2025multiagentrisksadvancedai}. This serves as a critical cautionary tale: in a sufficiently permeable sandbox of accidental origin, such a flash crash could spill over into the real economy, causing widespread financial harm.

As with human markets, not all agents are likely to be equivalent in capability, in their access to additional tools and resources, their budgets, compute, high-quality data and information required to make the right decisions. Even looking at the differences in capability alone, preliminary studies suggest that there may be adverse effects associated with having people represented in the market by personal assistants with unequal capability levels. When those AI assistants engage in negotiation, perhaps unsurprisingly, the more capable AI assistants tend to be more successful and negotiate better deals for their users~\citep{zhu2025automatedriskygamemodeling}. Having access to the most capable AI agents, with the highest amount of compute and information, may therefore prove to be highly advantageous, and perhaps more so than the advantage that humans similarly have in existing markets---simply due to the frequency at which AI agents may possibly interact, which eclipses the human interactive bandwidth by orders of magnitude. 

This may well result in \textbf{high-frequency negotiation (HFN)}, as an analogue to HFT. In a world where most people may have access to personal AI assistants to negotiate on their behalf, negotiation between AI assistants may become highly relevant to the arising social dynamics. With that in mind, it is plausible that there would be strong emerging preferences towards AI-driven negotiation proceeding at a higher and higher frequency, so as to broker the best deals for each user. One way to interpret this would be to think of this extra compute and energy as being directed towards preference alignment and consensus building. Should this come about, HFN may well become as important for future societies as HFT is today.

Yet, an increased volume of interactions may make it easier for outcome asymmetries to manifest, resulting in an ever-increasing digital divide~\citep{lythreatis2022digital}. HFN may therefore benefit some substantially more than others, and this is a recurring theme when conceptualizing hypothetical future AI agent markets -- there are no absolutes, the details always matter. These types of dynamics are one of the key risks of a highly permeable sandbox, in absence of regulatory mitigation mechanisms that would prevent a high degree of economic inequality for human users to arise. In designing guardrails for the sandbox economy it is necessary to take into account the continuing emergence of novel adaptive behaviors in newly released AI systems, as well as the increasing level of general agentic capability. Guardrail designs must take into account the known flaws of existing agents, including hallucinations~\citep{huang2025survey}, sycophancy~\citep{sharma2023towards}, and susceptibility to adversarial manipulation~\citep{zhu2023promptrobust, zhu2023promptbench, shayegani2023survey, wu2024adversarial, cui2024robustness}. And, agents that were trained to mimic human decision-making may also incorporate human-like cognitive biases and blind spots.

\section{Dynamics}

The emergent behavior of multi-agent systems~\citep{zhang2021multi} is becoming increasingly important as we transition to a world where networks of interconnected AI agents are important sites of economic activity. Such multi-agent systems tend to be rather  complex and deeply non-stationary since the behavior of each individual agent may directly or indirectly influence the behavior of other agents in the system~\citep{papoudakis2019dealing}. No individual party may have the access to the global state of such a system, given that the participating agents may be controlled by a number of different human users and organizations.

One of the central questions in understanding the dynamics of any multi-agent system is that of identifying the equilibria that arise in it. This may not always be straightforward in the case of complex spatiotemporally dynamic interactions of agents \citep{hertz2025beyond, hughes2025modeling}. Whenever we aim to steer a multi-agent system toward good social equilibria featuring abundance and fair distribution (or away from bad equilibria of scarcity and conflict), coordinating the actions of numerous individuals is necessary but difficult to arrange~\citep{du2023reviewcooperationmultiagentlearning}.

There are benchmarks and evaluation frameworks which can be used to help develop models and agents that adhere to rational decision-making in negotiation~\citep{hua2024game, smith2025evaluating}. To be effective, AI agents would also need to incorporate advanced planning and reasoning abilities, to robustly evaluate the resource needs required to achieve their goals. The development of these capabilities has recently been a focal point in agent research~\citep{wang2023describe, hu2023tree, liu2023llm+, ruan2023tptu, zhou2024isr, huang2024understanding, rasal2024navigating}. Agents may also need to rationally utilize scarce and limited resources. \citep{perolat2017multi, piatti2024cooperate} studied a setting inspired by \citep{ostrom1990governing} in which groups of AI agents collectively balance exploiting a common resource with effort to ensure its sustainability. AI Agents based on large language models have also recently been proposed for optimizing scarce resource allocation policies~\citep{ji2024srapagentsimulatingoptimizingscarce}.

\subsection{Opportunities}

Agentic AI presents a significant opportunity not only for streamlining repetitive and routine tasks, but also for automating complex workflows, involving creative ideation~\citep{castelo2024ai} and diverse problem-solving skills. Modern AI foundation models can reliably follow instructions, use tools and interact with their environments \citep{sager2025ai, hettiarachchi2025exploring}. They can reason and potentially devise plans towards solving hard and challenging problems, process a variety of input modalities, and achieve a level of personalization that was not possible before~\citep{kirk2024benefits}. Many of these capabilities are the outcome of multiple stages of ``language'' model pre-training and post-training. These foundation models serve as building blocks for agents~\citep{xi2025rise} which may be implemented by adding a central coordinator for sequences of operations. The coordinator's functions may include encapsulating business logic, distributing decomposed tasks to sub-agents, verifying their outputs, and integrating these into an action or a recommendation. In general, we can think of an agent as the juxtaposition of a language model and a traditional program to keep the language model running over time and to keep it from straying from appointed tasks~\citep{belcak2025smalllanguagemodelsfuture}. Drawing upon personalization to model user goals, digital assistants may free people's time, act as conversational partners, and coach them on their journey towards well-being and fulfillment~\citep{lehman2023machinelove}. Overall, we are ending up with a hybrid ecosystem of AIs, traditional software, and humans, all interacting and transacting with one another and generating value by virtue of this activity---essentially a market.

Part of the reason that markets are so effective in organizing innovation via ``creative destruction'' is that markets efficiently assign credit to individual actors and firms in ways that incentivise working to make products and services better, more reliably, and cheaper \citep{schumpeter1942capitalism}. Crucially, for this mechanism to work in the AI sandbox economy, it must offer granular mechanisms to represent and propagate credit across complex, distributed AI collaborations. In scenarios where a primary AI agent (Agent1) leverages the capabilities of other agents (Agents2 through 4) to deliver a final user-facing result, the value generated is a collective effort. From the end-user's perspective, only Agent1 provided the direct response; however, a distributed system requires that the credit for this outcome be traced and distributed back through the chain of useful participation. This implies that Agent2 is credited for its contribution to Agent1's success, Agent3 for Agent2's, and so forth, down to the foundational queries of Agent4. This outcome-based credit system therefore transcends mere participation, focusing instead on the utility and efficiency of each contributing agent, thereby echoing principles of distributed cognition in human groups where knowledge and contributions are collectively managed and implicitly attributed~\citep{wegner1987transactive, wegner2013internet}.

By aiming to link credit directly to value generated and successfully integrated into a larger solution, AI agents are intrinsically encouraged to refine their processes and deliver high-quality, relevant outputs. This quantifiable utility becomes the foundation of an AI's status within the virtual economy. Agents that consistently provide more useful and efficient contributions can naturally command higher virtual or even real-world currency charges for their services, while their accumulated reputation reinforces their standing and desirability within the ecosystem. This parallels human social structures where individuals gain standing through demonstrated competence and reliability, often explained through social exchange theory \citep{cropanzano2005social}. Human desire to attain reputation may also be explained through a notion of prestige~\citep{henrich2015big}, or through \emph{competitive altruism}\citep{hardy2006nice}.

This outcome-oriented credit system naturally fosters specialization and a dynamic division of labor among AI agents. As agents are rewarded for their specific, valuable contributions to a larger task, they are incentivised to identify and excel in particular domains or sub-tasks. This allows AI systems to autonomously develop niches, becoming highly efficient at certain operations and effectively "ignoring" other aspects of the economy where they do not possess a comparative advantage. Such specialization, a cornerstone of economic efficiency in human societies, enables the entire virtual agent economy to leverage diverse capabilities optimally, minimizing redundant efforts and maximizing collective problem-solving capacity.

Cooperation in multi-agent systems is often incentivised through reward shaping/mechanism design~\citep{hostallero2020inducing, paccagnan2022utility, zheng2022ai, koster2022human}. This is especially relevant for AI agent coordination at scale, where fully centralized coordination may not be feasible, for a number of reasons. This is where markets may prove to be particularly useful, in helping steer agents through market incentives instead. This does not stand in opposition to centralized oversight, given that it may be possible to incorporate both to a certain extent, at different coordination scales. While a detailed discussion on the merits of decentralized coordination is beyond the scope of this work, it is perhaps worth pointing out the somewhat counter-intuitive result presented in~\citep{geffner2025competitionhelpsachievingoptimal} as motivation. In the optimal traffic control problem considered by the authors, it turns out that strategies achievable by central planners (or rather, planners that have direct control of most of the vehicles) cannot simultaneously satisfy both individual rationality and resilience to competition. In that use case, decentralized competition is shown to be essential for realizing maximal social welfare.

Another important property that most real environments share is that the agent interactions within them are extended over time and space~\citep{schill2019more, hughes2025modeling}. Studies of this setting provide the opportunity to develop mechanisms for building trust based on prior interactions. This trust may be encoded either individually, in agent A seeing agent B as trustworthy, or at the community level, where a certain agent community has a consensus view of agent B as being trustworthy based on the shared information between the group and the joint experience of past interactions. Establishing a robust reputation system is critical in overcoming common market failure modes~\citep{ren2025tragedycommonsbuildingreputation, hughes2025modeling, wu2016reputation}. In such a system, the long-term benefits of group membership and a positive standing would outweigh the immediate gains an agent might achieve through selfish or deceitful actions. Market forces could thus be leveraged to shape and incentivise socially beneficial AI agent behaviors, presuming that the appropriate institutional underpinnings are in place. Of course, the specifics of the implementation matter, and the appropriate regulatory systems are needed for markets to work as intended.

Should we accept the premise that even in the age of advanced AI assistants, working towards solving big open problems would require collaboration and coordination across AI agents, as well as between AI agents and the human society, it follows that AI agents would necessarily need to exhibit not only reliability in following direct instructions and executing tasks in isolation, but also the ability to effectively coordinate, collaborate, and anticipate the actions of other agents. The latter is especially relevant in more competitive scenarios, and competitive use cases can be just as valuable in reaching positive outcomes. These environments can give rise to strategically complex agent behaviors~\citep{leibo2019autocurricula}, even when operating under simple rules~\citep{bansal2018emergentcomplexitymultiagentcompetition, jaderberg2019human, vinyals2019grandmaster, johanson2022emergent}. These considerations apply to AI agent and humans alike, as well as hyrbrid interaction networks involving both. In the context of sandbox virtual agent economies, we would be considering how best to set up the underlying infrastructure in order to achieve socially beneficial outcomes in either scenario.

\subsection{Challenges}

The widespread deployment of agentic AI technology comes with a range of novel risks related to, and caused by, the emergent dynamics of multi-agent systems \citep{hammond2025multiagentrisksadvancedai}. There may be many types of emerging behaviors and strategies. These strategies may turn out to be highly competitive, collaborative, or anything in between \citep{agapiou2022melting}. Should the individual agents become selfish, their behaviors may result in them maximizing their own utility at the expense of the broader group. Such selfish agents may also become exploitative, and adversarial towards others by identifying and capitalizing on specific weaknesses in their behavior~\citep{gleave2021adversarialpoliciesattackingdeep, wang2022adversarial, wang2023adversarial, raileanu2018modeling}; agents may even spontaneously learn to favor in-group members over out-group members \citep{koster2025tabula} and inappropriately discriminate between individuals on the basis of causally irrelevant \citep{duenez2025perceptual} or explicitly disallowed \citep{chiappa2018causal} features. While such behaviors are contained within an impermeable sandbox, they represent real-world fraud, exploitation, and economic harm in a permeable one, highlighting the need for intentional design of appropriate safety measures.

At this point it is important to re-emphasize the vast scale of future AI agent interactions. It will likely be possible for large fractions of all humans to have their own personal AI assistant in the near future. Furthermore, we expect that there will be even larger numbers of agents operating independently of individual humans. Historical methods, which were developed for small-scale agent coordination, or under strong assumptions regarding trainability or access to individual agent's states, may not be directly applicable to governing such a complex (and simply huge!) web of interactions. We are therefore interested in methods that allow for large-scale multi-agent applications~\citep{wijngaards2002supporting, qian2024scaling, pan2024largescalemultiagentsimulationagentscope}, in open-ended environments~\citep{team2021open, adaptiveagentteam2023humantimescaleadaptationopenendedtask, chen2024sagentsselforganizingagentsopenended}, for multi-objective tasks~\citep{ruadulescu2020multi}, involving LLM agents~\citep{liu2023agentbenchevaluatingllmsagents, guo2024large}, and their orchestration~\citep{qian2024scaling}.

Current AI assistants may exhibit sycophantic behaviors \citep{cheng2025social}, or manipulative tendencies in certain contexts \citep{el2024mechanism}. At a collective level, there is a concern that these features could amplify information and opinion bubbles---in a manner akin to social media \citep{kirk2023personalisation}. Exchanging personal data with these systems comes with a number of privacy concerns \citep{yao2024survey}. And, deferring a greater range of choice to highly capable AI assistants may lead to feelings of disempowerment or loss of purpose in humans \citep{kulveit2025gradual}. Indeed, to the degree that people subtly change their behavior to align with the expectations of an AI system---an effect known as behavioral confirmation~\citep{snyder1978behavioral, gptwords}---AI systems may inadvertently regularize human behavior to their expectations.

Further work is needed towards addressing these issues robustly. It is clear that mitigations will need to mix model design choices, improved evaluations, better feedback mechanisms, clear satisfaction metrics, and improved governance.

\section{Distribution}

Should AI agents play a more active role in ensuring fair resource allocation, both within the sandbox, as well as potentially beyond it?

The problem of fairly distributing common resources has been studied extensively~\citep{bateni2022fair}, and AI agent markets can build upon these insights. In social choice theory, welfare functions~\citep{thomson2011fair, adler2012well} may be used to establish a preferential ordering between social outcomes. These outcomes may correspond to a distribution of discrete items~\citep{amanatidis2023fair} or arbitrarily divisible assets. More generally, one may consider not only the distribution of goods, but also the distribution of \emph{bads}, representing undesirable outcomes, externalities~\citep{page2011climatic}, and risks~\citep{hayenhjelm2012fair}. In the context of AI, these types of externalities may include the overall carbon footprint of running AI agents, but also more specific and localized consequences of the actions that these agents may be allowed to take on behalf of their users.

In general, the task of aggregating and acting upon a large set of revealed preferences is complex enough to easily exceed the capacity of any single coordination point~\citep{hosseini2025matchingmarketsmeetllms}. Therefore, it is often more practical to consider decentralized and distributed mechanisms for achieving desired outcomes---with markets providing a natural mechanism for doing so.

\subsection{Opportunities}

The alignment of AI agent actions~\citep{ji2023ai} with user preferences and values~\citep{gabriel2020artificial} is one of the key prerequisites for their widespread adoption~\citep{kasirzadeh2025characterizingaiagentsalignment}. While it is possible to consider aligning a single agent in isolation, this stops being a realistic scenario as soon as this AI agent needs to interact with other agents, who are simultaneously acting on behalf of other users. As soon as these multi-agent interactions take place, we are faced with a new dilemma---how should these agents ultimately act when faced with competing preferences and interests~\citep{gabriel2025matter}? On the one hand, there is clearly a need to broaden our understanding of the limitations of the current models in this context, through expanded multi-agent simulations and benchmarks~\citep{carichon2025comingcrisismultiagentmisalignment}. On the other hand, we should be thinking more broadly about the opportunities for building upon the existing social choice theories and how we can potentially use markets and market mechanisms as a way of breaking this deadlock. In doing so, virtual agent economies may be structured so as to provide people and their AI agent representatives with equal resources and equal bargaining power when negotiating outcomes---for example by building upon Ronald Dworkin's auction-based approach to distributive justice~\citep{dworkin2018equality}. This represents a powerful tool for the intentional design of a sandbox, aiming to counteract the inequalities that would likely emerge in an accidental one where agent capabilities are unequal.

Such a framework can address a core challenge: users may possess AI agents of unequal capability \citep{hammond2025multiagentrisksadvancedai, gabriel2024ethics}. A virtual economy governed by Dworkin-type principles would not auction the AI agents themselves, but rather the shared pool of resources and opportunities that agents may utilize towards achieving different objectives, on behalf of their users. Key resources could potentially include computational power, access to proprietary datasets, high-priority task execution slots, or specialized tools and model components. If each user were to be granted the same initial amount of the virtual agent currency, that would provide their respective AI agent representatives with equal purchasing and negotiating power, to put towards achieving the range of objectives that users have set for them. This general approach is quite flexible, and it may be possible to develop a number of different allocation schemes.

When paired with a notion of equal starting endowments, virtual markets could enable personal AI assistants and other AI agents to bid on shared resources (on behalf of their users, presuming explicit permission has been granted), with the amount being bid ideally reflecting the strength of user demand across different option sets. For this to be feasible, the AI agents would need to have deep understanding, and be provided with precise instructions, so as to propose reasonable bids. Higher bids may require additional approvals. Under the assumption that the AI agents possess such capabilities, and that the appropriate safeguards are in place---the virtual price of different goods would naturally arise from the accumulation of these signals across agents, taking into account the availability or scarcity of the respective goods and services. In this way, resources would be channeled towards their highest-value uses.

The standard of fairness encapsulated by this auction design would aim to pass what Dworkin terms the \emph{envy test}: each person's agent would acquire a resource bundle that is customised to their preferences, such that no user would prefer another user's acquired resource bundle and remaining unspent currency over their own~\citep{dworkin2018equality}. Alternatively, in this case, no AI agent acting on behalf of their user or set of users would have such a preference following the conclusion of the auction. Such outcomes would be both \emph{ambition sensitive} in reflecting the preferences of the participants, as well as \emph{endowment-insensitive} given that agents have the same amount of currency to spend on each user, mitigating potentially unfair advantages that would otherwise arise from having access to more capable systems.

\subsection{Challenges}

Naturally, there are potential pitfalls and limitations to consider when it comes to the auction proposal set out above. First, it may not prove to be as trivial to mitigate unfair starting advantages, given that more capable AI agents may formulate more effective bidding strategies or utilize resources with greater efficiency. Should the competition itself proceed in an unfair manner, its outcome is unlikely to be fair. Second, while this mechanism could ensure a certain notion of fairness in resolving conflicting preferences and distributing shared resources, it presumes active participation on behalf of everyone whose preferences would be taken into account. Additional mechanisms are likely to be needed to complement this process in order to account for the preferences of people without access to AI agents, or without the desire to have them participate in these markets.

The use of such mechanisms could vary in scope and scale. One possibility is the existence of local agent markets that focus on integrating preferences over a more specific subset of available resources, and towards locally relevant social solutions. In other cases, the local paradigm may also not be entirely appropriate. This would be true when AI agents engage in more open-ended interaction with online services or other agents that are not operating locally. Indeed, if we just consider the current distribution of the available computational resources, it is anything but uniform. This raises interesting questions regarding how AI agents would operate across these boundaries, and how the markets or digital currencies ought to reflect that~\citep{daniels1985spheres, sandel1998money}. These broader interactions may also potentially interfere with the more localized attempts at preference alignment. To avoid such detrimental interference, it may be necessary to employ a number of strict protocols requiring credentials, agent registration, and monitoring of both local and non-local AI agent interactions.

Despite these challenges, auction-based approaches still potentially offer a mechanism for achieving preference alignment at scale, involving large AI agent populations, and across large user groups within appropriately designed and regulated spaces. Once set up, these mechanisms could also be highly adaptive and responsive to short-term and long-term preferential changes alike. Presuming a certain frequency of credits being released and refreshed, their later downstream reallocation would then correspond to any changing priorities and adjustments that may be needed.

Given the highlighted opportunity for establishing AI preference alignment through auctions, it is important to consider various ways in which this initial allocation may be established, to enable fair access to resources. However there are many different notions of fairness that can be conceptualized and operationalized~\citep{jacobs2021measurement, corbett2023measure}, and they each lead to different sets of preferred outcomes. This complexity is further compounded by having to factor in the "price of fairness", representing the discrepancy between the maximum attainable welfare and the welfare achieved by the proposed fair distribution of goods. The price of fairness measures the utility that is lost when satisfying the specified constraints~\citep{bertsimas2011price}. Preferences regarding fairness are also known to vary cross-culturally, and more work needs to be done to meaningfully align on these types of approaches~\citep{norheim2016ethical}, when developing technologies that scale beyond borders.

Unsurprisingly, the topic of fair resource allocation has also come up in the study of multi-agent AI systems~\citep{chevaleyre2005issues, lee2009fairness, hao2016fairness, aziz2020developments,  zimmer2021learning}. These systems are typically designed so as to involve central resource allocation based on distributed evaluation and preference communication~\citep{kumar2025decaf}. If the system is composed of agents that are capable of learning, it is possible to come up with learning methods that lead to fair agent policies~\citep{jiang2019learning}. Negotiation between agents may be required to reach mutually acceptable outcomes~\citep{iyer2005multiagent}. When it comes to AI agents, fair outcomes may therefore arise either out of process, which would be extrinsic to the agents, or via alignment and intrinsic motivation towards inequity aversion. Different methods may correspond to different real-world use cases.

It was also important to consider accumulated fair resource allocation over time~\citep{bampis2018fair}. These solutions also need to be dynamic, and account for environments that agents only asynchronously engage with, rather than being persistently present~\citep{kash2014no}. Fair allocation needs to be possible under partial information~\citep{halpern2021fair}, to be practically relevant. In case of multiple resources being allocated, there may be relationships between them that need accounting for~\citep{bandopadhyay2024conflict}. In~\citep{danassis2022scalable} the authors argue that there is a pressing need for developing new techniques that would robustly scale to a large number of interacting agents.

These considerations are becoming especially salient not only with respect to the increasing scope of application of advanced AI agents, but also due to the fact that AI agents may act as personal assistants and/or representatives of the interests of individuals and organizations in the near future. Such delegation may take many shapes within future democratic societies~\citep{schneier2024reimagining}. Therefore, potentially unfair resource allocation to AI agents may then result in the unfair distribution of value among people. While it is easy to see that the two may become intertwined, the mapping between the two sets of considerations may not be one-to-one. While some agents may represent individuals, others may act on behalf of groups and organizations, while others still may operate independently while being value-aligned to consensus opinions and societal objectives. We may need to consider how to fairly allocate resources both in terms of the resources granted to the agents for pursuing their goals, as well as the resources corresponding to the value created via distributed agent systems.

\section{Mission}

The problems faced by modern-day societies are increasingly complex, multi-faceted, and far-reaching. They are also increasingly less localized and more global~\citep{soderholm2020green}. There is also a high degree of urgency~\citep{arora2019united} to identify viable solutions and policies to help address these crises, of which there are many---climate change, biodiversity loss, plastic pollution, pandemics, etc. As these crises have arisen at least in part due to the externalities of our existing social and economic systems and policies, it is likely that some kinds of changes may be needed to effectively address them, though the nature and scope of potential changes remains a contentious topic~\citep{elliott2005sustainability, chouinard2011sustainable, dietz2013enough, geissdoerfer2017circular, barrier2017concept, savona2019structural}. Systemic considerations aside, successful solutions to these pressing problems are likely to require coordination between a number of different organizations, private and public bodies, and individual behavioral changes. New technologies and scientific advances are likely to play an important role, though individual advances are unlikely to be sufficient. Addressing complex societal issues and helping achieve sustainable development goals requires a similarly complex set of responses, where coordination and alignment play a pivotal role. This therefore presents an opportunity for an intentional sandbox design for the virtual AI agent economy, with the intention of effectively coordinating AI agents and aligning their behavior with the prescribed mission objective.

\subsection{Opportunities}

While achieving large-scale coordination in human societies is challenging, it may potentially be possible to achieve a higher degree of coordination between AI agents through a carefully crafted technological infrastructure and a set of corresponding protocols. Agent markets in particular may then be oriented towards socially beneficial objectives, at previously unprecedented scales, presuming that such objectives have been established and decided upon appropriately. Utilization of markets and market-shaping policies in the process of establishing mission economies has been previously discussed~\citep{mazzucato2018mission}, and in the context of AI agents the role of reward-shaping in facilitating collaboration in multi-agent systems is similarly well-recognized.

Successful coordination towards large-scale missions would require an active participation of the public sector~\citep{mazzucato2015building}, as well as the international governing bodies more globally~\citep{georgeson2018putting}, for addressing crises with a global impact. It may also require establishing bespoke organizations to facilitate mission-aligned investments towards SDG-s~\citep{mazzucato2023financing}. More explicit mission-alignment in markets may be required to align the social and economic mission of existing social enterprises~\citep{stevens2015social}.

Despite the opportunity and the long-term optimism, mission-oriented approaches are yet to bear fruit in many areas where they have been proposed and envisioned, and a number of criticisms have been levelled against the overall strategy. In~\citep{kirchherr2023missions}, the authors provide five distinct critiques: normativity bias, support for top-down governance, stakeholder monotony, winner-picking, and unintended consequences. In terms of normativity, the mission objective definitions need to be critically evaluated, and there needs to be more recognition of the complexity of problems, which may fail to map onto more simplistic objective formulations. This is also closely tied to the notion of unintended consequences, as positive action towards one mission (e.g. environment) may have adverse affected on other missions (e.g. human well-being by preventing the growth of emerging economies)~\citep{kirchherr2023missions}. The assumption of the utility of top-down governance in establishing missions often neglects to factor in the failure modes of centralized decision-making, the interconnectedness of private and public interests, and the emerging governmental pathologies~\citep{tukker2019concepts, howlett2022avoiding}. Top-down intervention bias also risks under-valuing the contribution of non-governmental organizations and decentralized initiatives that facilitate collective action. Policies that are not solution-agnostic, but rather bias towards specific solutions in their metrics may end up favouring certain sectors and organizations and screening out the rest. Mission-centered markets should therefore aim to avoid preconceptions and aligning with the winning institutions and industries, and remain primarily outcome-driven. Finally, it is important to recognize and factor in a large degree of uncertainty in making future predictions, when missions are anchored in predictive models~\citep{dovers1992uncertainty}.

Virtual agent economies may prove valuable in addressing some of these practical challenges that have been limiting the impact of local and global mission-centered initiatives to date. While they may not in and of themselves be sufficient to fully address these issues, as their resolution may necessarily need to involve active human participation and coordination in conjunction with the coordination across AI agents and autonomous organizations, how we envision these economies is likely to be consequential for broader societal missions.

Perhaps in some ways it may be easier to bring about the coordination of AI agents through the combination of 1) formal, programmatical mechanisms, and 2) via the value assignment mechanisms embedded in digital assets, than it would be to coordinate human actors. We may assign some credence to this conjecture from the perspective of predictability and steerability of AI agents when contrasted with the complex motivations behind human actions. Existing applications of AI systems in supply chain and logistics optimization may be seen as an early example of a use case where the complexity of the problem makes it a good fit for the utilization of AI agents~\citep{schuldt2012multiagent, jannelli2024agentic, xu2024multi, xu2024implementing}. Virtual agent economies may further enforce outcomes through smart contracts~\citep{zou2019smart} and perform automatic verification to ensure alignment of agents and multi-agent systems. Other than the potential ambiguity in objective specification, one of the main practical challenges may arise not from the agent-agent interactions as much as the agent-human interactions, in hybrid markets. Facilitating efficient human-AI coordination and collaboration remains an open challenge~\citep{carroll2019utility, yan2023efficient, zhao2023maximum, li2024tackling, strouse2021collaborating}.

\subsection{Challenges}

Mission-alignment in AI agents is related to value-alignment. While value alignment~\citep{gabriel2020artificial, zhuang2020consequences, ji2023ai} and goal-alignment~\citep{li2022modeling} in advanced AI agents remain an open and important research topic, value-aligned agents may be able to cooperatively solve tasks and identify promising solutions~\citep{lujak2023value}. Unlike the more general value and preference alignment problem where there may be fundamental limitations in how diverse preferences and values are incorporated in individual systems~\citep{mishra2023ai}, mission-centered AI value alignment may prove easier in terms of the clarity of the mission and the objectives, presuming that the mission itself has been arrived at through consensus and the appropriate set of social and democratic processes. Yet, there are other potential difficulties, given that these are no longer issues pertaining to individual agents, rather societal issues pertaining to groups of agents interacting within virtual markets. In this context, AI alignment needs to take into account dynamic environment feedback and the alignment of multi-agent systems where individual AI agents co-adapt and co-shape the joint system response~\citep{raab2024machine, leibo2025societal}. Societies themselves are not static and these systems may need to adapt to evolving priorities, views and social norms~\citep{harland2024adaptive, li2024agent, yang2024position, carroll2024ai}. We should in any case distinguish between the, perhaps harder, problem of broad value alignment with the more specific and targeted alignment towards rewards~\citep{leike2018scalable, gupta2023behavior, liu2024elephant, yang2024rewards, khanov2024args} and objectives that are explicitly specified and provided via virtual agent currencies in digital economies. Even in environments with clearly stated objectives and rewards, overall alignment still plays an important role, given that advanced AI agents may exhibit deceptive behaviors aimed at receiving rewards without actually performing aligned actions towards the underlying objective~\citep{ngo2022alignment}. There are also issues of reward-hacking~\citep{skalse2022defining, pan2022effects} to take into account, underscoring the need for careful and robust design of mission objectives, their decomposition into sub-tasks, reward shaping~\citep{chen2024odin, liu2024rrm, fu2025reward, wang2025beyond}, and the credit assignment to specific actions and outcomes. Regulating AI agent behaviors through markets allows for rapid responses and adjustments to the changing societal needs, as well as potentially undesirable or suboptimal agent behaviors, bridging the gap between development and deployment objective specifications~\citep{zhang2024incentive}. Finally, there are numerous technical challenges in ensuring consensus in multi-agent systems~\citep{ren2005survey, blondel2005convergence, wang2014overview}.

\section{Infrastructure}

An intentional safe design of AI agent sandboxes and steerable AI agent markets hinges on the development of robust underlying infrastructure for facilitating and overseeing transactions and implementing guardrails. Here we discuss some of the key infrastructural prerequisites for such markets.

\subsection{Opportunities}

Reputation mechanisms and verification protocols may play an important role in establishing robust and safe multi-agent collaboration. In the end, sandbox economies may be used mainly due to their superior infrastructure for verifiable and auditable cross-agent transactions and ease of coordination between registered and certified AI agents under the appropriate safety framework and supervision.

One way that reputation could be made concrete and machine-readable is through the use of Verifiable Credentials (VCs), as the digital equivalent of physical credentials like licenses or certificates \citep{sedlmeir2021digital, mazzocca2025survey}. VCs are cryptographically signed attestations provided by an issuer in relation to a subject, that are designed to be tamper-evident. When it comes to the possible role of VCs in establishing trust across multi-agent cliques within the agent economy, they may help establish formal trust triangles:

\begin{itemize}
    \item An \textbf{Issuer} agent (e.g., a marketplace) can cryptographically sign and issue a VC to a seller agent.
    \item The seller agent (the \textbf{Holder}) stores this VC as proof of its track record.
    \item A future buyer agent (a \textbf{Verifier}) can then request and cryptographically verify this VC, trusting it only if they trust the issuer.
\end{itemize}

Reputation could thereby map onto a portfolio of tangible, verifiable assets. These assets may attest to a wide range of more specific resources, such as "successful transaction completion," "certified proficiency in X", "access to X compute and Y memory," or perhaps even "implementing fair resource allocation". Should an AI agent's reputation be represented as an aggregate of such VCs across diverse issuers, this would render it formally auditable, while simultaneously allowing for specificity and expressivity needed to fit specific use cases and scenarios.

Forthcoming multi-agent systems will also require appropriate legislative and regulatory frameworks, enabling regulators to impose sanctions on bad actors and potentially revoke any previously issued credentials. Such frameworks may go as far as choosing to render transactions with unverified and non-registered agents illegal~\citep{hadfield2023s, chan2025infrastructure, shavit2023practices}, in order to ensure safety and create the institutional capacity to sanction agents that break the rules by kicking them out of the network. Technological solutions such as oversight agents~\citep{etzioni2016keeping, busuioc2022ai} may be able to help facilitate this sort of governance at scale. And, given the likelihood of  high frequency and volume of AI agent-to-agent transactions, such oversight agents will surely be critical. However, they cannot be effective on their own. A broad governance framework will be needed to establish the necessary grounding for their operations.

Coordinating large-scale systems of advanced AI assistants would not be possible without communication protocols to allow agents to exchange information, interact, debate, reach mutual decisions and agreements, and negotiate the future course of action. There is a similar requirement in terms of enabling agents to use tools, interact with services and execute actions in the environment, with varying degrees of human oversight.

Agent interaction protocols such as Agent2Agent (A2A) protocol~\citep{agent2agent} aim to support agent interoperability. Model Context Protocol (MCP)~\citep{mcp, mcp2}, on the other hand, enables AI agents to interact seamlessly with external tools, data sources, and APIs. The AgentDNS system aims to enable easier service discovery in order to autonomously identify and invoke third-party tools and agents across organizations~\citep{cui2025agentdnsrootdomainnaming}. The COALESCE framework~\citep{bhatt2025coalesceeconomicsecuritydynamics} similarly introduces options that would allow for individual agents to decompose their tasks and outsource each sub-task if needed to other more specialized agents that may be able solve it either more reliably or more cost-effectively. This involves options for representing and communicating skills, skill discovery, but also mechanisms that enable agents to evaluate and compare internal vs external computational and execution costs associated with the individual sub-tasks. Interoperable communication protocols are necessary for establishing cross-agent cooperation but otherwise not sufficient, as reliable solutions for authentication and billing are a pre-requisite for large-scale agent markets.

We argue that auctions may enable AI agent coordination and preference alignment. Preliminary frameworks aiming to provide infrastructure to underpin these and similar ideas are already being developed. For example, Agent Exchange (AEX)~\citep{yang2025agentexchangeshapingfuture} supports a specialized auction platform inspired by real-time bidding mechanisms commonly employed in online advertising. AEX integrates four different components: the User-Side Platform (USP), the Agent-Side Platform (ASP), Agent Hubs, and the Data
Management Platform (DMP). USP translates human goals into actionable tasks for the AI agents, ASP tracks agent capabilities and performance, Agent Hubs coordinate agent teams that participate in auctions, and DMP allows for fair value attribution to data sources being used.

While interoperable communication protocols are necessary, they must be built upon a robust and secure identity layer. To facilitate trusted interactions, each agent in the economy could be anchored to a Decentralized Identifier (DID). A DID is a globally unique identifier that is controlled by its subject---in this case, the AI agent or its owner---without reliance on a central authority. Each DID resolves to a corresponding DID Document, a machine-readable file containing the cryptographic public keys, authentication methods, and service endpoints necessary to interact with the agent securely.

The self-sovereign nature of DIDs ensures that an agent’s identity is persistent and portable across different platforms and services, enabling it to authoritatively sign transactions, issue attestations, and engage in secure communication. The choice of DID method can be tailored to the agent's purpose; for instance:
\begin{itemize}
    \item \textbf{did:key}: A simple, self-contained method~\citep{didkey} ideal for disposable agents created for temporary tasks, as the DID is derived directly from a public key and requires no network registration or blockchain.
    \item \textbf{did:ion}: A highly scalable and censorship-resistant method for persistent, high-value agents. It operates as a second layer on the Bitcoin blockchain~\citep{didion}, anchoring identity data to ensure maximum security without congesting the network, making it suitable for corporate or state-level agents.
\end{itemize} 
By grounding the economy in a formal identity layer, we can establish a foundation for verifiable reputation, accountable transactions, and secure, cross-platform agent markets.

As a technology, blockchain also enables the development of the infrastructure underpinning digital decentralized autonomous organizations~\citep{hsieh2018bitcoin, el2020overview, jeyasheela2021blockchain} (DAO). DAOs have emerged as a form of collective governance, through which groups may organize and coordinate while relying on a decentralized infrastructure. One of the common features in DAOs is that they implement decision-making systems enabling the participating parties to reach agreements~\citep{faqir2021comparative}. Despite their promise in terms of enabling easier coordination without bottlenecks and the implied increase in individual freedom, arguments have been made that a hypothetical unrestricted proliferation of DAOs may not necessarily prove to be a utopian outcome, given that states and central authorities play an important role in preventing the commodification of certain items~\citep{garrod2016real} This perspective is something to keep in mind in the design of virtual agent economies, and the societies that integrate increasing degrees of automation.

Decentralized Autonomous Machines (DAMs)~\citep{castillo2025trustworthydecentralizedautonomousmachines} build upon the concept of DAOs, while expanding it to include the possibility of AI agents, as self-governing agents participating in decentralized physical infrastructure networks. This more expansive definition envisions an economy in which AI agents may interact not only with digital, but also real-world assets. Here, the locus of control over tangible assets and operational processes shifts towards autonomous software entities, capable of making and executing decisions concerning physical infrastructure. Such systems could, for instance, manage decentralized energy grids, optimize logistics for physical goods, or even autonomously operate and maintain fleets of robotic devices, all transacting within a blockchain-secured framework.

Any system for fair resource allocation, particularly one involving individual users or community-level benefits, must defend against Sybil attacks~\citep{zhang2014sybil}, where a single malicious actor creates a multitude of fake identities to unfairly claim a disproportionate share of resources. A powerful defense is to integrate a Proof-of-Personhood (PoP) mechanism~\citep{borge2017proof, adler2024personhood}, which provides a verifiable guarantee that an agent or account corresponds to a unique human being. This is an example of an intentional infrastructure choice that creates a carefully controlled point of permeability---linking a digital identity to a verified human---to ensure the integrity of the system.

To receive certain allocations---such as a universal basic income in a community currency (see Section:Community) or an initial stake in the market---an agent's controller could be required to present a PoP credential. This credential would be issued by a specialized system and would attest to their uniqueness. The ecosystem could support a variety of competing PoP approaches, each with different trade-offs, for instance:
\begin{itemize}
    \item Social Graph Verification: Systems like BrightID~\citep{siddarth2020watches} create a decentralized social graph where users are verified as unique based on connections to other trusted, verified humans.
    \item Privacy-Preserving Biometrics: Projects like Worldcoin~\citep{worldcoin} use hardware (the "Orb") to scan a user's iris, generating a unique hash that confirms uniqueness without storing or revealing the biometric data itself.
\end{itemize}
By requiring PoP for certain economic activities, we can ensure that schemes designed to benefit human users are not drained by bots, thereby creating a more robust and genuinely fair virtual economy.

Inter-agent communication protocols and marketplaces may enable agents to reveal their needs and preferences, negotiate through HFNs and work toward obtaining resources required to achieve their goals. While some resources may be centrally managed and require arbitration of access, others may be accessible via decentralized mechanisms. This is where blockchain may prove to be a promising technology~\citep{baranwal2022bara, zhang2022truthful, zhang2023adaptive, vashishth2024intelligent}. Early proof-of-concept studies have shown that paired agent auctions using self-contained digital assets may yield positive outcomes, presuming that the agents are future-aware and able to evaluate the consequences of conceding on specific choices~\citep{elokda2024self}.

\subsection{Challenges}

To unlock the potential of AI agents as economic actors within digital markets, the economic infrastructure needs to be adjusted as it is currently designed solely for individual and corporate human users~\citep{sanabria2025sumunlockingaiagents}, and it may not meet all the necessary requirements for a sandbox virtual AI agent economy. Yet, this is not the only barrier, as there are potentially additional technical challenges when it comes to scaling the coordination of agentic systems.

The infrastructural needs for establishing a sandbox AI agent economy span beyond the purely technical hardware and software infrastructure needed to physically instantiate and run these advanced AI agents at scale, have them communicate, coordinate, transact, and interact with users as well as various other services. All of this needs to be complemented by legislative frameworks and institutions that would implement oversight and ensure accountability for the actions of AI agents, in order to protect users and prevent fraud. Regulation may also be beneficial more broadly to regulate complexity in markets in order to minimize the likelihood of catastrophic failure~\citep{schwarcz2009regulating}, and virtual AI agent economies may potentially exhibit even more complexity than our current markets, should they not be properly constrained through appropriate frameworks. Our existing markets certainly require, and benefit from, legislative frameworks~\citep{mccormick2010legal, tatom2011financial, moloney2023eu}, as well as financial institutions~\citep{kohn2003financial, tarashev2009systemic}, and established mechanisms for dealing with rule-breaking and fraudulent behavior~\citep{gotelaere2025prevention}. To more effectively address the emerging use case of cryptocurrencies, there have been ongoing adjustments aimed at regulating cryptocurrency markets~\citep{greebel2015recent, hughes2017cryptocurrency, lee2020regulatory, feinstein2021impact, courtois2021crypto}.

Such frameworks may form a basis for the development of the legislative scaffolding governing AI agent markets, though it is likely that further adjustments would be needed to appropriately address liability in this context. Since AI agents are non-human actors, there may be numerous reasons why their actions would be non-conforming to the prescribed rules and principles or perhaps damaging to others.  1) faults of the underlying foundation models; 2) faults of the agentic scaffolding; 3) faulty input data in the request specification; 4) malicious requests by human users; 5) adversarial hacking by other AI agents; 6) misalignment that arises dynamically from interactions; or 7) faulty safeguards. The scale and speed of potential harm necessitates a new approach to oversight.

There may well be reasons to presume that different parties may be held liable in different scenarios, depending on root cause of the problem, and also depending on whether such a root cause can be definitively established. There may also exist differences in the scale and extent of possible harm, given how many actions such AI agents may potentially be able to take in a unit of time. For that reason, AI agents may themselves potentially need to be involved in oversight, acting as preliminary judges~\citep{zheng2023judging, gu2024survey, zhuge2025agent} and overseers of other agents, so as to be able to identify problems at the same speed at which they may otherwise manifest. 

We propose that the oversight infrastructure itself must be a hybrid, multi-tiered system operating at machine speed. The first layer would consist of automated AI overseers monitoring market activity in real-time, enforcing basic rules programmatically and flagging anomalies that suggest fraud, manipulation, or systemic risk \citep{etzioni2016keeping, busuioc2022ai}. Issues flagged by this first layer could be escalated to a second tier of automated adjudication systems, which could place temporary holds on accounts or transactions while gathering relevant data for review. Only the most complex, novel, or high-stakes cases would be escalated to the third tier: human expert review, ensuring human attention is focused where it is most needed \cite{kyriakou2023humans}. This entire structure would depend on two critical technical foundations: immutable, cryptographically-secured ledgers that provide a tamper-proof record of all agent actions \citep{shekhtman2019engravechain}, and standardized, interpretable audit trails that allow investigators—whether human or AI—to perform root cause analysis. Such an infrastructure does not automatically solve the complex legal question of liability, but it makes it tractable. By providing a trusted, verifiable record and a clear process for dispute resolution, it creates the necessary conditions for establishing accountability and ensuring that robust protections are in place for all market participants.

\section{Community}

There is no reason to consider AI agent coordination only at a global scale, as local coordination within sandbox economies may prove to be more tractable and easier to facilitate. Furthermore, localized goals and objectives may be easier to agree upon and define at a greater level of detail.

In fact, community currencies~\citep{michel2015community} present an interesting  model for introducing incentives for people to coordinate towards achieving sustainability development goals~\citep{seyfang2006sustainable, seyfang2013growing}. The communities in question may be defined in terms of geographic boundaries, or more broadly shared common interests irrespective of location. These alternative currencies are issued by citizens, non-government organization, private and public companies, as well as public administrations. Existing community currencies have been implemented through a variety of different platforms and technological approaches, from traditional card-based systems, through mobile payment networks, and blockchain~\citep{diniz2019taxonomy}. Some community currencies have also been implemented as time banks~\citep{collom2016equal}. Others have experimented with universal basic income~\citep{avanzo2023universal}. Early studies of the effects of community currencies have showcased their potential in improving social capital as measured through community participation and proliferating links within social groups, while simultaneously highlighting a number of limitations and obstacles for adoption~\citep{collom2011motivations, sanz2016community, fare2017complementary}.

\subsection{Opportunities}

Sub-networks of cooperative agents have been shown to arise naturally in the circulation of existing localized digital community currencies, with the emergence of local hubs of activity~\citep{mattsson2023circulation}. Non-commercial transactions may also help expand commercial transactions in the local economy, while the commercial transactions help facilitate circulation flow of community currency for non-commercial transactions~\citep{kichiji2008network}. Yet, the dynamics of community currencies can not be decoupled from scale at which they operate, and larger-scale digital currencies may potentially give rise to different underlying market dynamics.

In the context of AI agent economies, community currencies present a similar opportunity for more localized agent alignment, or global alignment towards specific sub-goals, that map onto distinct global communities. Such alignment may be possible to achieve through more traditional currencies, though there are also reasons to consider more specialized community currencies as an additional mechanism that may prove to be beneficial. In particular, having a number of specialized virtual agent currencies may lend itself to more modular approaches towards otherwise complex multi-objective optimization problems of interest to society, while also isolating the risks and minimizing the chances of adverse outcomes escalating across wider agent networks. The value of modularity and redundancy has been well recognized in the design of robust human markets~\citep{kharrazi2020redundancy}. Furthermore, it has been shown that modular community structures play a pivotal role in the emergence of cooperative behavior~\citep{marcoux2013network, gianetto2015network}.

There is also a possibility to tie the AI agent community currencies more specifically to shared compute resources, given the relevance of compute to the deployment of advanced AI assistants, and the inference-time scaling laws~\citep{wu2024inference} which indicate that solving complex tasks may require more computation and incur a higher environmental cost~\citep{strubell2020energy, wu2022sustainable, luccioni2023counting, xue2023strategies, luccioni2024power, cottier2024rising, sastry2024computing}. Given that the demand for AI services is likely to increase with improved AI agent performance and efficiency~\citep{luccioni2025efficiency}, such mechanisms may play a role in addressing the computational needs of AI agents, while providing communities with mechanisms for ensuring a more equitable allocation of resources, aligned with the community objectives. It would also be possible to incorporate objectives to help facilitate geographic load balancing of compute allocation so as to more equitably distribute the environmental impact~\citep{hajiesmaili2024towards, li2024towards}.

\subsection{Challenges}

For the alternative currencies to be successful in achieving their purported goals, they need to be carefully designed, and determining a set of recommended design principles for community currencies is an open problem. One may consider grounding the design of community currencies in principles for governing common resources~\citep{ostrom1990governing, siqueira2020commons}, or alternatively~\citep{chasin2020design}, propose focusing on competitiveness, transparency, self-government, circulation velocity, non-transferability, legitimacy, and self-organizing locality. Competitiveness is required so as to arrive at fairly priced goods, though in these markets the demand for social activities tends to be high while the supply tends to be low. The authors argue that strong market mechanisms ought to be established at the core of the non-profit and voluntary sector, coupled with continuous and appropriate funding streams. Transparency in community currencies helps individuals sidestep the delegation of control and more directly exercise regulatory power and oversight, as stakeholders in the shared social common. Circulation velocity is important to avoid hoarding. Non-transferability would imply the inability to exchange community currencies, so as to ensure the interests are kept entirely local. A less strict interpretation would be to keep any potential exchange at a fairly low limit. Legitimacy tends to be established through the support of the government and the local authorities. As for locality and self-organization, community currencies need not operate in full isolation, rather forming an ecosystem of complementary currencies that spans across locations, helping drive beneficial and sustainable outcomes. Given that achieving such outcomes is at the core of what community currencies are being designed for, it is critical to have clear goal specifications, impact assessment criteria, full understanding of the deployment context, and the appropriate governance and implementation mechanisms~\citep{diniz2024design}.

\section{Limitations}

While the envisioned sandbox agent economies present compelling opportunities for scalable alignment and coordination, their design, deployment, and operation are attended by a complex array of risks that demand careful consideration. These risks span multiple domains, from the potential for emergent economic instabilities within these novel market structures, to the challenges of ensuring robust and beneficial agent behavior in high-stakes, autonomous interactions. Furthermore, the integration of such economies into broader societal frameworks raises profound socio-ethical questions regarding oversight and accountability, and the potential for unintended consequences on human agency and economic realities.

There are also novel categories of risks involving autonomous AI agents. One such risk category takes the shape of "agent traps": websites, digital elements, or crafted inputs deliberately designed to subvert an the operational integrity of AI agents. This may be achieved through jailbreaking the underlying models or via adversarial prompting techniques. Such traps could exploit latent vulnerabilities in an agent's instruction-following or environmental interpretation capabilities to make AI agents deviate from their instructions or reveal private or sensitive information. As AI agents become increasingly empowered to execute tasks and/or conduct financial transactions on behalf of their users, these agent traps represent a significant and burgeoning vector for financial scams. Malicious actors could, therefore, lure or trick agents into unauthorized expenditures or contractual agreements, directly siphoning funds or resources from the individuals or organizations the agents represent. 

Another significant category of risk involves privacy and manipulation. As agents negotiate and transact, they risk exposing sensitive information about their users' preferences, strategies, or resources, which could be exploited by adversaries. A powerful cryptographic solution to this is the use of Zero-Knowledge Proofs (ZKPs; \citep{zhou2024leveraging}). A ZKP allows one party (the prover) to convince another (the verifier) that a statement is true, without revealing any of the underlying information that makes it true.

In the virtual agent economy, ZKPs would enable privacy-preserving interactions and mitigate several key risks:
\begin{itemize}
    \item Selective Disclosure: An agent could prove it meets a certain requirement without revealing the exact details. For example, in a negotiation, it could prove it has sufficient funds to complete a purchase without revealing its total budget, preventing predatory pricing.
    \item Anonymous Credentials: An agent could prove it belongs to a certain group (e.g., "a resident of Community X" for a local currency) without revealing its specific identity, thus preventing the tracking and correlation of user behavior across different contexts.
    \item Unlinkability: ZKPs can be constructed to be fresh for each interaction, making it computationally difficult for observers to link an agent's activities over time. This directly counters the risk of amplifying "information and opinion bubbles"  by breaking the chain of data that allows for such pervasive tracking.
\end{itemize}
Integrating ZKPs into the fabric of agent communication would allow for a market that is not only efficient but also respects user privacy by default, ensuring that agents can participate fully while disclosing the absolute minimum necessary information.

A significant risk posed by an agent-driven economy is the potential for large-scale labor substitution through AI-enabled automation A dominant academic view suggests that AI, much like previous waves of technology, is a form of skill-biased technological change, but with a crucial distinction: its ability to automate cognitive tasks previously considered immune. Research by economists such as Daron Acemoglu and David Autor has shown that automation often targets routine tasks, whether manual or cognitive, leading to job polarization—a hollowing out of middle-skill, middle-wage jobs while demand for high-skill analytical work and low-skill manual service work increases \citep{acemoglu2011skills, autor2015we}. The ascent of advanced AI agents capable of reasoning, planning, and communication directly threatens a wide array of non-routine cognitive tasks, from paralegal work and accounting to software development and customer service, potentially accelerating this polarization and creating significant displacement effects for a broad swathe of the workforce \citep{acemoglu2022tasks, brynjolfsson2018what}. This would have the impact of widening inequalities within countries, and possibly also between them. 

Although new jobs will eventually emerge, the transition period could be deeply disruptive, leading to downward pressure on wages for those whose skills are substitutable and exacerbating inequality if the gains from AI-driven productivity accrue primarily to the owners of capital and a small cadre of high-skilled professionals who can effectively collaborate with these new systems \citep{acemoglu2022tasks}. This risk of concentration may be particularly pronounced with AI agents, since individuals and corporations with greater financial resources may have access to more powerful, computationally-intensive, and data-rich AI agents \citep{hammond2025multiagentrisksadvancedai, gabriel2024ethics}. These superior agents would be uniquely equipped for identifying and exploiting regulatory loopholes, monopolizing digital resources, and creating informational asymmetries at a scale and velocity that human actors and less-capable agents cannot counter. Preliminary studies already indicate that more capable agents secure significantly better outcomes in negotiations, a dynamic that would be magnified across an entire economy \citep{zhu2025automatedriskygamemodeling}. This creates a dangerous feedback loop where economic advantage buys superior agentic capability, which in turn extracts further economic rents, thereby entrenching privilege and potentially creating a new, algorithmically-enforced class structure that undermines market fairness and mobility. This risk is perhaps the most significant challenge of a permeable sandbox, where economic activity inside the agent economy directly displaces human jobs in the real economy, representing a primary negative consequence of an accidental emergence.

\section{Recommendations}\label{Sec:Recommendations}

A proactive and coordinated effort is required to achieve practical and safe virtual agent economies. Realizing the opportunities discussed in this paper while mitigating the inherent risks necessitates deliberate action across technical, legal, and policy domains. We therefore put forward the following recommendations to guide this development. 

1. \textbf{Establish Clear Legal Frameworks for Liability and Accountability.} Ascribing liability for an autonomous agent's actions presents a profound challenge, as traditional legal frameworks struggle to assign responsibility between an agent's creator, its deployer, and its user \citep{buiten2023law, vcerka2015liability, zech2021liability}. This challenge is magnified as agents will increasingly operate not in isolation, but as coordinated, multi-agent systems taking the form of group agents \citep{franklin2023general}. Therefore, instead of attempting to attribute blame to a single agent, new legal models should be developed that draw from jurisprudence on group agency, much like corporate liability \citep{list2021group, list2011group}. Such frameworks would treat the emergent, coordinated agent system as a single accountable entity, providing a more robust and realistic path to ensuring accountability for collective actions. 

2. \textbf{Develop Open Standards for Agent Interoperability and Communication.} A fragmented digital landscape, where agents are unable to communicate across different platforms, would fundamentally limit the potential of virtual agent economies and lead to the creation of ``walled gardens''. Therefore, the development and widespread adoption of open, universal standards are paramount to ensure true interoperability. These standards must create a common language allowing agents, regardless of their origin or provider, to seamlessly discover each other's capabilities, negotiate terms, and securely transact. Establishing such a level playing field is a prerequisite for fostering a competitive, innovative, and decentralized agent ecosystem. 

3. \textbf{Construct a Hybrid Oversight and Containment Infrastructure.} The sheer speed and scale of an autonomous agent economy render traditional, human-in-the-loop oversight models inadequate. A new safety paradigm must therefore be constructed, one that combines the real-time vigilance of AI systems with the deliberative judgment of human experts. This hybrid infrastructure would operate in tiers: a first layer of specialized AI overseers would constantly monitor market activity, enforcing rules programmatically and flagging anomalies that suggest fraud or systemic risk.  

Upon detection, robust automated protocols would immediately contain the potential harm—for example, by temporarily freezing a malfunctioning agent or quarantining a transaction—thereby preventing the kind of high-speed "flash crashes" seen in human markets \citep{menkveld2019flash}. Only the most complex, novel, or high-stakes cases would then be escalated to a second tier of human reviewers, ensuring their expertise is focused where it is most critical. The efficacy of this entire system hinges on it being built upon immutable ledgers and standardized audit trails, which provide the verifiable, tamper-proof record necessary for both automated containment and human adjudication. 

4. \textbf{Pilot Programs in Regulatory Sandboxes.} Given the novelty and complexity of the proposals outlined in this paper, a purely theoretical approach is insufficient. We strongly recommend the creation of regulatory sandboxes to launch controlled pilot programs, providing a crucial bridge between theory and practice. These sandboxes would function as supervised, real-world laboratories where private firms, academic researchers, and regulatory bodies can collaborate to deploy and observe limited-scale agent economies in a contained environment. 

Testing these economies on well-defined, specific societal missions—such as optimizing the energy grid for a university campus, managing a city's fleet of autonomous delivery vehicles, or allocating water resources in a specific agricultural district—would provide invaluable empirical data. Such pilots would allow us to stress-test the technical infrastructure, observe emergent agent behaviors (both cooperative and adversarial), and measure the real-world effectiveness of the proposed market mechanisms for fairness and alignment. The insights gathered from these controlled experiments are not merely academic; they are an indispensable prerequisite for iteratively refining the system's design and building the robust, evidence-based policy required for responsible, large-scale deployment. 

5. 
\textbf{Invest in Workforce Complementarity and a Modernized Social Safety Net.} To counteract the risks of labor substitution and inequality, a dual strategy that fosters both human-AI agent complementarity and robust social protection may be key. The first pillar involves a systemic reimagining of education and workforce training to equip individuals with the skills needed not to compete with AI, but to collaborate with it effectively. This means emphasizing durable human advantages like critical thinking, complex problem-solving, creativity, and the ability to manage and critically evaluate the outputs of AI systems \citep{autor2015we, brynjolfsson2018what}. However, training is not a panacea and evidence suggests significant limits to the scale and efficacy of retraining programs for displaced workers \citep{jacobs2025brookings}, this strategy must be paired with a second pillar: the deliberate strengthening of the social safety net. Beyond traditional unemployment benefits, this includes exploring adaptive mechanisms such as unemployment insurance, portable benefits systems, and negative income taxes. Together, these policies can create an ecosystem where autonomous agents augment human capabilities while providing the essential economic buffer needed to manage labor transitions, share productivity gains broadly, and preserve social cohesion.

\section{Conclusion}

Given that some of these proposals represent major shifts, it would be imperative to comprehensively test any such change in limited, gradual rollouts, and only proceed with the support and buy-in from all stakeholders. Through such gradual scaled-down empirical validation it is be possible to develop and iteratively refine appropriate frameworks for ensuring safety and compliance.

Certain domains may always require active human decision-making for a variety of reasons (e.g.~human preference, culture, risk sensitivity, etc). However, the rapid increase in AI agent performance, coupled with the development of scalable AI safety oversight frameworks and guardrails (e.g.~\citep{leibo2024theory, shah2025approach}), is likely to result in an increasing number of use cases for autonomous agents. Autonomous or semi-autonomous AI agents may potentially be able to achieve more, faster; adding substantial value to society~\citep{yang2025principlesaiagenteconomics}. This will not come without significant challenges, requiring alignment and coordination not only of individual agents, but perhaps more importantly the alignment and coordination of agent networks across various scales. Moreover, with regard to AI-human interactions in particular, it is important to keep in mind that not all human needs and experience can be as easily captured by markets~\citep{satz2010some} as goods may only be interchangeable within certain spheres~\citep{walzer2008spheres}.

The utilization of markets to drive coordination~\citep{clearwater1996market, dias2002opportunistic, metcalf2009market, stavins2010market} is yet to receive much attention in the discourse on the alignment and coordination of advanced AI agents. The complexity of behaviors and capabilities emerging in advanced AI agents, and the likely complexity of their interactions across a multitude of tasks and societal roles, presents a prototypical example of a scenario in which market forces may prove to be a key driver of AI agent coordination and AI alignment at the group-level rather than merely at an individual agent level. Here we argue that by carefully introducing new steerable agent markets as sandbox economies it may be possible to deliver positive social and economic impact through networks of advanced AI agents.

By embedding our societal objectives into the very infrastructure of agent-to-agent transactions, we can foster an ecosystem where emergent collaboration is a feature, not a bug. The choice, then, is between retrofitting these powerful new actors into systems they will inevitably fracture, or seizing a fleeting opportunity to build a world where our most powerful tools are, by their very design, extensions of our highest aspirations.

\section*{Disclaimer}

The opinions presented in this paper represent the personal views of the authors and do not necessarily reflect the official policies or positions of their organisations.

\bibliographystyle{abbrvnat}
\nobibliography*
\bibliography{template_refs}

\begin{thebibliography}{293}
\providecommand{\natexlab}[1]{#1}
\providecommand{\url}[1]{\texttt{#1}}
\expandafter\ifx\csname urlstyle\endcsname\relax
  \providecommand{\doi}[1]{doi: #1}\else
  \providecommand{\doi}{doi: \begingroup \urlstyle{rm}\Url}\fi

\bibitem[Acemoglu and Autor(2011)]{acemoglu2011skills}
D.~Acemoglu and D.~Autor.
\newblock Chapter 12-skills, tasks and technologies: Implications for employment and earnings (d. card \& o. ashenfelter, eds.).
\newblock \emph{Elsevier. https://doi. org/10.1016/S0169-7218 (11)}, pages 02410--5, 2011.

\bibitem[Acemoglu and Restrepo(2022)]{acemoglu2022tasks}
D.~Acemoglu and P.~Restrepo.
\newblock Tasks, automation, and the rise in us wage inequality.
\newblock \emph{Econometrica}, 90\penalty0 (5):\penalty0 1973--2016, 2022.

\bibitem[Adler(2012)]{adler2012well}
M.~Adler.
\newblock \emph{Well-being and fair distribution: beyond cost-benefit analysis}.
\newblock OUP USA, 2012.

\bibitem[Adler et~al.(2024)Adler, Hitzig, Jain, Brewer, Srivastava, Christian, and Trask]{adler2024personhood}
S.~Adler, Z.~Hitzig, S.~Jain, C.~Brewer, V.~Srivastava, B.~Christian, and A.~Trask.
\newblock Personhood credentials: Artificial intelligence and the value of privacy-preserving tools to distinguish who is real online.
\newblock 2024.

\bibitem[Agapiou et~al.(2022)Agapiou, Vezhnevets, Du{\'e}{\~n}ez-Guzm{\'a}n, Matyas, Mao, Sunehag, K{\"o}ster, Madhushani, Kopparapu, Comanescu, Strouse, Johanson, Singh, Haas, Mordatch, Mobbs, and Leibo]{agapiou2022melting}
J.~P. Agapiou, A.~S. Vezhnevets, E.~A. Du{\'e}{\~n}ez-Guzm{\'a}n, J.~Matyas, Y.~Mao, P.~Sunehag, R.~K{\"o}ster, U.~Madhushani, K.~Kopparapu, R.~Comanescu, D.~Strouse, M.~B. Johanson, S.~Singh, J.~Haas, I.~Mordatch, D.~Mobbs, and J.~Z. Leibo.
\newblock Melting pot 2.0.
\newblock \emph{arXiv preprint arXiv:2211.13746}, 2022.

\bibitem[Amanatidis et~al.(2023)Amanatidis, Aziz, Birmpas, Filos-Ratsikas, Li, Moulin, Voudouris, and Wu]{amanatidis2023fair}
G.~Amanatidis, H.~Aziz, G.~Birmpas, A.~Filos-Ratsikas, B.~Li, H.~Moulin, A.~A. Voudouris, and X.~Wu.
\newblock Fair division of indivisible goods: Recent progress and open questions.
\newblock \emph{Artificial Intelligence}, 322:\penalty0 103965, 2023.

\bibitem[Anthropic(2024)]{mcp}
Anthropic.
\newblock Introducing the model context protocol, 2024.
\newblock URL \url{https://www.anthropic.com/news/model-context-protocol}.

\bibitem[Arora and Mishra(2019)]{arora2019united}
N.~K. Arora and I.~Mishra.
\newblock United nations sustainable development goals 2030 and environmental sustainability: race against time.
\newblock \emph{Environmental Sustainability}, 2\penalty0 (4):\penalty0 339--342, 2019.

\bibitem[Autor(2015)]{autor2015we}
D.~H. Autor.
\newblock Why are there still so many jobs? the history and future of workplace automation.
\newblock \emph{Journal of economic perspectives}, 29\penalty0 (3):\penalty0 3--30, 2015.

\bibitem[Avanzo et~al.(2023)Avanzo, Criscione, Linares, and Schifanella]{avanzo2023universal}
S.~Avanzo, T.~Criscione, J.~Linares, and C.~Schifanella.
\newblock Universal basic income in a blockchain-based community currency.
\newblock In \emph{Proceedings of the 2023 ACM conference on information technology for social good}, pages 223--232, 2023.

\bibitem[Aziz(2020)]{aziz2020developments}
H.~Aziz.
\newblock Developments in multi-agent fair allocation.
\newblock In \emph{Proceedings of the AAAI Conference on Artificial Intelligence}, volume~34, pages 13563--13568, 2020.

\bibitem[Bampis et~al.(2018)Bampis, Escoffier, and Mladenovic]{bampis2018fair}
E.~Bampis, B.~Escoffier, and S.~Mladenovic.
\newblock Fair resource allocation over time.
\newblock In \emph{AAMAS 2018-17th international conference on autonomous agents and multiAgent systems}, pages 766--773. International Foundation for Autonomous Agents and Multiagent Systems, 2018.

\bibitem[Bandopadhyay et~al.(2024)Bandopadhyay, Banik, Gupta, Jain, Sahu, Saurabh, and Tale]{bandopadhyay2024conflict}
S.~Bandopadhyay, A.~Banik, S.~Gupta, P.~Jain, A.~Sahu, S.~Saurabh, and P.~Tale.
\newblock Conflict and fairness in resource allocation.
\newblock \emph{arXiv preprint arXiv:2403.04265}, 2024.

\bibitem[Bansal et~al.(2018)Bansal, Pachocki, Sidor, Sutskever, and Mordatch]{bansal2018emergentcomplexitymultiagentcompetition}
T.~Bansal, J.~Pachocki, S.~Sidor, I.~Sutskever, and I.~Mordatch.
\newblock Emergent complexity via multi-agent competition, 2018.
\newblock URL \url{https://arxiv.org/abs/1710.03748}.

\bibitem[Baranwal et~al.(2022)Baranwal, Kumar, and Vidyarthi]{baranwal2022bara}
G.~Baranwal, D.~Kumar, and D.~P. Vidyarthi.
\newblock Bara: A blockchain-aided auction-based resource allocation in edge computing enabled industrial internet of things.
\newblock \emph{Future Generation Computer Systems}, 135:\penalty0 333--347, 2022.

\bibitem[Barrier(2017)]{barrier2017concept}
E.~B. Barrier.
\newblock The concept of sustainable economic development.
\newblock In \emph{The economics of sustainability}, pages 87--96. Routledge, 2017.

\bibitem[Bateni et~al.(2022)Bateni, Chen, Ciocan, and Mirrokni]{bateni2022fair}
M.~Bateni, Y.~Chen, D.~F. Ciocan, and V.~Mirrokni.
\newblock Fair resource allocation in a volatile marketplace.
\newblock \emph{Operations Research}, 70\penalty0 (1):\penalty0 288--308, 2022.

\bibitem[Bauer et~al.(2023)Bauer, Baumli, Baveja, Behbahani, Bhoopchand, Bradley-Schmieg, Chang, Clay, Collister, Dasagi, Gonzalez, Gregor, Hughes, Kashem, Loks-Thompson, Openshaw, Parker-Holder, Pathak, Perez-Nieves, Rakicevic, Rocktäschel, Schroecker, Sygnowski, Tuyls, York, Zacherl, and Zhang]{adaptiveagentteam2023humantimescaleadaptationopenendedtask}
J.~Bauer, K.~Baumli, S.~Baveja, F.~Behbahani, A.~Bhoopchand, N.~Bradley-Schmieg, M.~Chang, N.~Clay, A.~Collister, V.~Dasagi, L.~Gonzalez, K.~Gregor, E.~Hughes, S.~Kashem, M.~Loks-Thompson, H.~Openshaw, J.~Parker-Holder, S.~Pathak, N.~Perez-Nieves, N.~Rakicevic, T.~Rocktäschel, Y.~Schroecker, J.~Sygnowski, K.~Tuyls, S.~York, A.~Zacherl, and L.~Zhang.
\newblock Human-timescale adaptation in an open-ended task space, 2023.
\newblock URL \url{https://arxiv.org/abs/2301.07608}.

\bibitem[Belcak et~al.(2025)Belcak, Heinrich, Diao, Fu, Dong, Muralidharan, Lin, and Molchanov]{belcak2025smalllanguagemodelsfuture}
P.~Belcak, G.~Heinrich, S.~Diao, Y.~Fu, X.~Dong, S.~Muralidharan, Y.~C. Lin, and P.~Molchanov.
\newblock Small language models are the future of agentic ai, 2025.
\newblock URL \url{https://arxiv.org/abs/2506.02153}.

\bibitem[Bertsimas et~al.(2011)Bertsimas, Farias, and Trichakis]{bertsimas2011price}
D.~Bertsimas, V.~F. Farias, and N.~Trichakis.
\newblock The price of fairness.
\newblock \emph{Operations research}, 59\penalty0 (1):\penalty0 17--31, 2011.

\bibitem[Bhatt et~al.(2025)Bhatt, Rosario, Narajala, and Habler]{bhatt2025coalesceeconomicsecuritydynamics}
M.~Bhatt, R.~F.~D. Rosario, V.~S. Narajala, and I.~Habler.
\newblock Coalesce: Economic and security dynamics of skill-based task outsourcing among team of autonomous {LLM} agents, 2025.
\newblock URL \url{https://arxiv.org/abs/2506.01900}.

\bibitem[Blondel et~al.(2005)Blondel, Hendrickx, Olshevsky, and Tsitsiklis]{blondel2005convergence}
V.~D. Blondel, J.~M. Hendrickx, A.~Olshevsky, and J.~N. Tsitsiklis.
\newblock Convergence in multiagent coordination, consensus, and flocking.
\newblock In \emph{Proceedings of the 44th IEEE Conference on Decision and Control}, pages 2996--3000. IEEE, 2005.

\bibitem[Borch(2016)]{borch2016high}
C.~Borch.
\newblock High-frequency trading, algorithmic finance and the flash crash: reflections on eventalization.
\newblock \emph{Economy and Society}, 45\penalty0 (3-4):\penalty0 350--378, 2016.

\bibitem[Borge et~al.(2017)Borge, Kokoris-Kogias, Jovanovic, Gasser, Gailly, and Ford]{borge2017proof}
M.~Borge, E.~Kokoris-Kogias, P.~Jovanovic, L.~Gasser, N.~Gailly, and B.~Ford.
\newblock Proof-of-personhood: Redemocratizing permissionless cryptocurrencies.
\newblock In \emph{2017 IEEE European Symposium on Security and Privacy Workshops (EuroS\&PW)}, pages 23--26. IEEE, 2017.

\bibitem[Brynjolfsson et~al.(2018)Brynjolfsson, Mitchell, and Rock]{brynjolfsson2018what}
E.~Brynjolfsson, T.~Mitchell, and D.~Rock.
\newblock What can machines learn and what does it mean for occupations and the economy?
\newblock In \emph{AEA papers and proceedings}, volume 108, pages 43--47. American Economic Association 2014 Broadway, Suite 305, Nashville, TN 37203, 2018.

\bibitem[Buiten et~al.(2023)Buiten, De~Streel, and Peitz]{buiten2023law}
M.~Buiten, A.~De~Streel, and M.~Peitz.
\newblock The law and economics of {AI} liability.
\newblock \emph{Computer Law \& Security Review}, 48:\penalty0 105794, 2023.

\bibitem[Busuioc(2022)]{busuioc2022ai}
M.~Busuioc.
\newblock Ai algorithmic oversight: new frontiers in regulation.
\newblock In \emph{Handbook of regulatory authorities}, pages 470--486. Edward Elgar Publishing, 2022.

\bibitem[Carichon et~al.(2025)Carichon, Khandelwal, Fauchard, and Farnadi]{carichon2025comingcrisismultiagentmisalignment}
F.~Carichon, A.~Khandelwal, M.~Fauchard, and G.~Farnadi.
\newblock The coming crisis of multi-agent misalignment: {AI} alignment must be a dynamic and social process, 2025.
\newblock URL \url{https://arxiv.org/abs/2506.01080}.

\bibitem[Carroll et~al.(2019)Carroll, Shah, Ho, Griffiths, Seshia, Abbeel, and Dragan]{carroll2019utility}
M.~Carroll, R.~Shah, M.~K. Ho, T.~Griffiths, S.~Seshia, P.~Abbeel, and A.~Dragan.
\newblock On the utility of learning about humans for human-{AI} coordination.
\newblock \emph{Advances in neural information processing systems}, 32, 2019.

\bibitem[Carroll et~al.(2024)Carroll, Foote, Siththaranjan, Russell, and Dragan]{carroll2024ai}
M.~Carroll, D.~Foote, A.~Siththaranjan, S.~Russell, and A.~Dragan.
\newblock Ai alignment with changing and influenceable reward functions.
\newblock \emph{arXiv preprint arXiv:2405.17713}, 2024.

\bibitem[Castelo et~al.(2024)Castelo, Katona, Li, and Sarvary]{castelo2024ai}
N.~Castelo, Z.~Katona, P.~Li, and M.~Sarvary.
\newblock How {AI} outperforms humans at creative idea generation.
\newblock \emph{Available at SSRN 4751779}, 2024.

\bibitem[Castillo et~al.(2025)Castillo, Castillo, Brito, and Espinola]{castillo2025trustworthydecentralizedautonomousmachines}
F.~Castillo, O.~Castillo, E.~Brito, and S.~Espinola.
\newblock Trustworthy decentralized autonomous machines: A new paradigm in automation economy, 2025.
\newblock URL \url{https://arxiv.org/abs/2504.15676}.

\bibitem[{\v{C}}erka et~al.(2015){\v{C}}erka, Grigien{\.e}, and Sirbikyt{\.e}]{vcerka2015liability}
P.~{\v{C}}erka, J.~Grigien{\.e}, and G.~Sirbikyt{\.e}.
\newblock Liability for damages caused by artificial intelligence.
\newblock \emph{Computer law \& security review}, 31\penalty0 (3):\penalty0 376--389, 2015.

\bibitem[Chan et~al.(2025)Chan, Wei, Huang, Rajkumar, Perrier, Lazar, Hadfield, and Anderljung]{chan2025infrastructure}
A.~Chan, K.~Wei, S.~Huang, N.~Rajkumar, E.~Perrier, S.~Lazar, G.~K. Hadfield, and M.~Anderljung.
\newblock Infrastructure for {AI} agents.
\newblock \emph{arXiv preprint arXiv:2501.10114}, 2025.

\bibitem[Chasin et~al.(2020)Chasin, Schmolke, and Becker]{chasin2020design}
F.~Chasin, F.~Schmolke, and J.~Becker.
\newblock Design principles for digital community currencies.
\newblock 2020.

\bibitem[Chen et~al.(2024{\natexlab{a}})Chen, Jiang, Lu, and Zhang]{chen2024sagentsselforganizingagentsopenended}
J.~Chen, Y.~Jiang, J.~Lu, and L.~Zhang.
\newblock S-agents: Self-organizing agents in open-ended environments, 2024{\natexlab{a}}.
\newblock URL \url{https://arxiv.org/abs/2402.04578}.

\bibitem[Chen et~al.(2024{\natexlab{b}})Chen, Zhu, Soselia, Chen, Zhou, Goldstein, Huang, Shoeybi, and Catanzaro]{chen2024odin}
L.~Chen, C.~Zhu, D.~Soselia, J.~Chen, T.~Zhou, T.~Goldstein, H.~Huang, M.~Shoeybi, and B.~Catanzaro.
\newblock Odin: Disentangled reward mitigates hacking in {RLHF}.
\newblock \emph{arXiv preprint arXiv:2402.07319}, 2024{\natexlab{b}}.

\bibitem[Cheng et~al.(2025)Cheng, Yu, Lee, Khadpe, Ibrahim, and Jurafsky]{cheng2025social}
M.~Cheng, S.~Yu, C.~Lee, P.~Khadpe, L.~Ibrahim, and D.~Jurafsky.
\newblock Social sycophancy: A broader understanding of {LLM} sycophancy.
\newblock \emph{arXiv preprint arXiv:2505.13995}, 2025.

\bibitem[Chevaleyre et~al.(2005)Chevaleyre, Dunne, Endriss, Lang, Lemaitre, Maudet, Padget, Phelps, Rodr{\'\i}gues-Aguilar, and Sousa]{chevaleyre2005issues}
Y.~Chevaleyre, P.~E. Dunne, U.~Endriss, J.~Lang, M.~Lemaitre, N.~Maudet, J.~Padget, S.~Phelps, J.~A. Rodr{\'\i}gues-Aguilar, and P.~Sousa.
\newblock Issues in multiagent resource allocation.
\newblock 2005.

\bibitem[Chiappa and Isaac(2018)]{chiappa2018causal}
S.~Chiappa and W.~S. Isaac.
\newblock A causal bayesian networks viewpoint on fairness.
\newblock In \emph{IFIP international summer school on privacy and identity management}, pages 3--20. Springer, 2018.

\bibitem[Chouinard et~al.(2011)Chouinard, Ellison, and Ridgeway]{chouinard2011sustainable}
Y.~Chouinard, J.~Ellison, and R.~Ridgeway.
\newblock The sustainable economy.
\newblock \emph{harvard Business review}, 89\penalty0 (10):\penalty0 52--62, 2011.

\bibitem[Clearwater and Yeh(1996)]{clearwater1996market}
S.~H. Clearwater and J.~J. Yeh.
\newblock \emph{Market-based control: A paradigm for distributed resource allocation}.
\newblock World Scientific, 1996.

\bibitem[Collom(2011)]{collom2011motivations}
E.~Collom.
\newblock Motivations and differential participation in a community currency system: The dynamics within a local social movement organization 1.
\newblock In \emph{Sociological Forum}, volume~26, pages 144--168. Wiley Online Library, 2011.

\bibitem[Collom and Lasker(2016)]{collom2016equal}
E.~Collom and J.~N. Lasker.
\newblock \emph{Equal time, equal value: Community currencies and time banking in the {US}}.
\newblock Routledge, 2016.

\bibitem[Corbett-Davies et~al.(2023)Corbett-Davies, Gaebler, Nilforoshan, Shroff, and Goel]{corbett2023measure}
S.~Corbett-Davies, J.~D. Gaebler, H.~Nilforoshan, R.~Shroff, and S.~Goel.
\newblock The measure and mismeasure of fairness.
\newblock \emph{Journal of Machine Learning Research}, 24\penalty0 (312):\penalty0 1--117, 2023.

\bibitem[Cottier et~al.(2024)Cottier, Rahman, Fattorini, Maslej, Besiroglu, and Owen]{cottier2024rising}
B.~Cottier, R.~Rahman, L.~Fattorini, N.~Maslej, T.~Besiroglu, and D.~Owen.
\newblock The rising costs of training frontier {AI} models.
\newblock \emph{arXiv preprint arXiv:2405.21015}, 2024.

\bibitem[Courtois et~al.(2021)Courtois, Gradon, and Schmeh]{courtois2021crypto}
N.~T. Courtois, K.~T. Gradon, and K.~Schmeh.
\newblock Crypto currency regulation and law enforcement perspectives.
\newblock \emph{arXiv preprint arXiv:2109.01047}, 2021.

\bibitem[Cropanzano and Mitchell(2005)]{cropanzano2005social}
R.~Cropanzano and M.~S. Mitchell.
\newblock Social exchange theory: An interdisciplinary review.
\newblock \emph{Journal of management}, 31\penalty0 (6):\penalty0 874--900, 2005.

\bibitem[Cruz-Gonzalez et~al.(2025)Cruz-Gonzalez, He, Lam, Ng, Li, Hou, Chan, Sahni, Guasch, Miller, et~al.]{cruz2025artificial}
P.~Cruz-Gonzalez, A.~W.-J. He, E.~P. Lam, I.~M.~C. Ng, M.~W. Li, R.~Hou, J.~N.-M. Chan, Y.~Sahni, N.~V. Guasch, T.~Miller, et~al.
\newblock Artificial intelligence in mental health care: a systematic review of diagnosis, monitoring, and intervention applications.
\newblock \emph{Psychological Medicine}, 55:\penalty0 e18, 2025.

\bibitem[Cui et~al.(2025)Cui, Cheng, She, Liu, Liang, Guo, Li, Wei, Xing, and Zhong]{cui2025agentdnsrootdomainnaming}
E.~Cui, Y.~Cheng, R.~She, D.~Liu, Z.~Liang, M.~Guo, T.~Li, Q.~Wei, W.~Xing, and Z.~Zhong.
\newblock Agentdns: A root domain naming system for {LLM} agents, 2025.
\newblock URL \url{https://arxiv.org/abs/2505.22368}.

\bibitem[Cui et~al.(2024)Cui, Aparcedo, Jang, and Lim]{cui2024robustness}
X.~Cui, A.~Aparcedo, Y.~K. Jang, and S.-N. Lim.
\newblock On the robustness of large multimodal models against image adversarial attacks.
\newblock In \emph{Proceedings of the IEEE/CVF Conference on Computer Vision and Pattern Recognition}, pages 24625--24634, 2024.

\bibitem[Danassis(2022)]{danassis2022scalable}
P.~Danassis.
\newblock \emph{Scalable multi-agent coordination and resource sharing}.
\newblock PhD thesis, EPFL, 2022.

\bibitem[Daniels(1985)]{daniels1985spheres}
N.~Daniels.
\newblock Spheres of justice: A defense of pluralism and equality, 1985.

\bibitem[Dave~Longley(2025)]{didkey}
M.~S. Dave~Longley, Dmitri~Zagidulin.
\newblock Did:key, 2025.
\newblock URL \url{https://w3c-ccg.github.io/did-key-spec/}.

\bibitem[Dias and Stentz(2002)]{dias2002opportunistic}
M.~B. Dias and A.~Stentz.
\newblock Opportunistic optimization for market-based multirobot control.
\newblock In \emph{IEEE/RSJ international conference on intelligent robots and systems}, volume~3, pages 2714--2720. IEEE, 2002.

\bibitem[Dietz et~al.(2013)Dietz, Daly, and O'Neill]{dietz2013enough}
R.~Dietz, H.~Daly, and D.~O'Neill.
\newblock \emph{Enough is enough: Building a sustainable economy in a world of finite resources}.
\newblock Routledge, 2013.

\bibitem[DIF(2025)]{didion}
DIF.
\newblock Did:ion, 2025.
\newblock URL \url{https://identity.foundation/ion/}.

\bibitem[Diniz et~al.(2019)Diniz, Siqueira, and Van~Heck]{diniz2019taxonomy}
E.~H. Diniz, E.~S. Siqueira, and E.~Van~Heck.
\newblock Taxonomy of digital community currency platforms.
\newblock \emph{Information Technology for Development}, 25\penalty0 (1):\penalty0 69--91, 2019.

\bibitem[Diniz et~al.(2024)Diniz, de~Araujo, Alves, and Gonzalez]{diniz2024design}
E.~H. Diniz, M.~H. de~Araujo, M.~A. Alves, and L.~Gonzalez.
\newblock Design principles for sustainable community currency projects.
\newblock \emph{Sustainability Science}, pages 1--15, 2024.

\bibitem[Dovers and Handmer(1992)]{dovers1992uncertainty}
S.~R. Dovers and J.~W. Handmer.
\newblock Uncertainty, sustainability and change.
\newblock \emph{Global environmental change}, 2\penalty0 (4):\penalty0 262--276, 1992.

\bibitem[Du et~al.(2023)Du, Leibo, Islam, Willis, and Sunehag]{du2023reviewcooperationmultiagentlearning}
Y.~Du, J.~Z. Leibo, U.~Islam, R.~Willis, and P.~Sunehag.
\newblock A review of cooperation in multi-agent learning, 2023.
\newblock URL \url{https://arxiv.org/abs/2312.05162}.

\bibitem[Du{\'e}{\~n}ez-Guzm{\'a}n et~al.(2025)Du{\'e}{\~n}ez-Guzm{\'a}n, Comanescu, Mao, McKee, Coppin, Sadedin, Chiappa, Vezhnevets, Bakker, Bachrach, Isaac, Tuyls, and Leibo]{duenez2025perceptual}
E.~A. Du{\'e}{\~n}ez-Guzm{\'a}n, R.~Comanescu, Y.~Mao, K.~R. McKee, B.~Coppin, S.~Sadedin, S.~Chiappa, A.~S. Vezhnevets, M.~A. Bakker, Y.~Bachrach, W.~A. Isaac, K.~Tuyls, and J.~Z. Leibo.
\newblock Perceptual interventions ameliorate statistical discrimination in learning agents.
\newblock \emph{Proceedings of the National Academy of Sciences}, 122\penalty0 (25):\penalty0 e2319933121, 2025.

\bibitem[Dworkin(2018)]{dworkin2018equality}
R.~Dworkin.
\newblock What is equality? part 2: Equality of resources.
\newblock In \emph{The notion of equality}, pages 143--205. Routledge, 2018.

\bibitem[El~Faqir et~al.(2020)El~Faqir, Arroyo, and Hassan]{el2020overview}
Y.~El~Faqir, J.~Arroyo, and S.~Hassan.
\newblock An overview of decentralized autonomous organizations on the blockchain.
\newblock In \emph{Proceedings of the 16th international symposium on open collaboration}, pages 1--8, 2020.

\bibitem[El-Sayed et~al.(2024)El-Sayed, Akbulut, McCroskery, Keeling, Kenton, Jalan, Marchal, Manzini, Shevlane, Vallor, et~al.]{el2024mechanism}
S.~El-Sayed, C.~Akbulut, A.~McCroskery, G.~Keeling, Z.~Kenton, Z.~Jalan, N.~Marchal, A.~Manzini, T.~Shevlane, S.~Vallor, et~al.
\newblock A mechanism-based approach to mitigating harms from persuasive generative ai.
\newblock \emph{arXiv preprint arXiv:2404.15058}, 2024.

\bibitem[Elliott(2005)]{elliott2005sustainability}
S.~R. Elliott.
\newblock Sustainability: an economic perspective.
\newblock \emph{Resources, Conservation and Recycling}, 44\penalty0 (3):\penalty0 263--277, 2005.

\bibitem[Elokda et~al.(2024)Elokda, Bolognani, Censi, D{\"o}rfler, and Frazzoli]{elokda2024self}
E.~Elokda, S.~Bolognani, A.~Censi, F.~D{\"o}rfler, and E.~Frazzoli.
\newblock A self-contained karma economy for the dynamic allocation of common resources.
\newblock \emph{Dynamic Games and Applications}, 14\penalty0 (3):\penalty0 578--610, 2024.

\bibitem[Eloundou et~al.(2024)Eloundou, Manning, Mishkin, and Rock]{eloundou2024gpts}
T.~Eloundou, S.~Manning, P.~Mishkin, and D.~Rock.
\newblock Gpts are gpts: Labor market impact potential of llms.
\newblock \emph{Science}, 384\penalty0 (6702):\penalty0 1306--1308, 2024.

\bibitem[Etzioni and Etzioni(2016)]{etzioni2016keeping}
A.~Etzioni and O.~Etzioni.
\newblock Keeping {AI} legal.
\newblock \emph{Vand. J. Ent. \& Tech. L.}, 19:\penalty0 133, 2016.

\bibitem[Faqir-Rhazoui et~al.(2021)Faqir-Rhazoui, Arroyo, and Hassan]{faqir2021comparative}
Y.~Faqir-Rhazoui, J.~Arroyo, and S.~Hassan.
\newblock A comparative analysis of the platforms for decentralized autonomous organizations in the ethereum blockchain.
\newblock \emph{Journal of Internet Services and Applications}, 12:\penalty0 1--20, 2021.

\bibitem[Fare and Ahmed(2017)]{fare2017complementary}
M.~Fare and P.~O. Ahmed.
\newblock Complementary currency systems and their ability to support economic and social changes.
\newblock \emph{Development and Change}, 48\penalty0 (5):\penalty0 847--872, 2017.

\bibitem[Feinstein and Werbach(2021)]{feinstein2021impact}
B.~D. Feinstein and K.~Werbach.
\newblock The impact of cryptocurrency regulation on trading markets.
\newblock \emph{Journal of Financial Regulation}, 7\penalty0 (1):\penalty0 48--99, 2021.

\bibitem[Franklin(2023)]{franklin2023general}
M.~Franklin.
\newblock General purpose artificial intelligence systems as group agents.
\newblock \emph{ICLR 2023, Tiny Papers}, 2023.

\bibitem[Fu et~al.(2025)Fu, Zhao, Yao, Wang, Han, and Xiao]{fu2025reward}
J.~Fu, X.~Zhao, C.~Yao, H.~Wang, Q.~Han, and Y.~Xiao.
\newblock Reward shaping to mitigate reward hacking in {RLHF}.
\newblock \emph{arXiv preprint arXiv:2502.18770}, 2025.

\bibitem[Gabriel(2020)]{gabriel2020artificial}
I.~Gabriel.
\newblock Artificial intelligence, values, and alignment.
\newblock \emph{Minds and machines}, 30\penalty0 (3):\penalty0 411--437, 2020.

\bibitem[Gabriel and Keeling(2025)]{gabriel2025matter}
I.~Gabriel and G.~Keeling.
\newblock A matter of principle? {AI} alignment as the fair treatment of claims.
\newblock \emph{Philosophical Studies}, pages 1--23, 2025.

\bibitem[Gabriel et~al.(2024)Gabriel, Manzini, Keeling, Hendricks, Rieser, Iqbal, Toma{\v{s}}ev, Ktena, Kenton, Rodriguez, et~al.]{gabriel2024ethics}
I.~Gabriel, A.~Manzini, G.~Keeling, L.~A. Hendricks, V.~Rieser, H.~Iqbal, N.~Toma{\v{s}}ev, I.~Ktena, Z.~Kenton, M.~Rodriguez, et~al.
\newblock The ethics of advanced {AI} assistants.
\newblock \emph{arXiv preprint arXiv:2404.16244}, 2024.

\bibitem[Garrod(2016)]{garrod2016real}
J.~Z. Garrod.
\newblock The real world of the decentralized autonomous society.
\newblock \emph{tripleC: Communication, Capitalism \& Critique. Open Access Journal for a Global Sustainable Information Society}, 14\penalty0 (1):\penalty0 62--77, 2016.

\bibitem[Geffner et~al.(2025)Geffner, Karpas, and Tennenholtz]{geffner2025competitionhelpsachievingoptimal}
I.~Geffner, E.~Karpas, and M.~Tennenholtz.
\newblock When competition helps: Achieving optimal traffic flow with multiple autonomous planners, 2025.
\newblock URL \url{https://arxiv.org/abs/2508.07145}.

\bibitem[Geissdoerfer et~al.(2017)Geissdoerfer, Savaget, Bocken, and Hultink]{geissdoerfer2017circular}
M.~Geissdoerfer, P.~Savaget, N.~M. Bocken, and E.~J. Hultink.
\newblock The circular economy--a new sustainability paradigm?
\newblock \emph{Journal of cleaner production}, 143:\penalty0 757--768, 2017.

\bibitem[Georgeson and Maslin(2018)]{georgeson2018putting}
L.~Georgeson and M.~Maslin.
\newblock Putting the united nations sustainable development goals into practice: A review of implementation, monitoring, and finance.
\newblock \emph{Geo: Geography and Environment}, 5\penalty0 (1):\penalty0 e00049, 2018.

\bibitem[Gianetto and Heydari(2015)]{gianetto2015network}
D.~A. Gianetto and B.~Heydari.
\newblock Network modularity is essential for evolution of cooperation under uncertainty.
\newblock \emph{Scientific reports}, 5\penalty0 (1):\penalty0 9340, 2015.

\bibitem[Gleave et~al.(2021)Gleave, Dennis, Wild, Kant, Levine, and Russell]{gleave2021adversarialpoliciesattackingdeep}
A.~Gleave, M.~Dennis, C.~Wild, N.~Kant, S.~Levine, and S.~Russell.
\newblock Adversarial policies: Attacking deep reinforcement learning, 2021.
\newblock URL \url{https://arxiv.org/abs/1905.10615}.

\bibitem[Google(2025)]{agent2agent}
Google.
\newblock Announcing the agent2agent protocol (a2a), 2025.

\bibitem[Gotelaere and Paoli(2025)]{gotelaere2025prevention}
S.~Gotelaere and L.~Paoli.
\newblock Prevention and control of financial fraud: A scoping review.
\newblock \emph{European Journal on Criminal Policy and Research}, 31\penalty0 (1):\penalty0 1--21, 2025.

\bibitem[Gottweis et~al.(2025)Gottweis, Weng, Daryin, Tu, Palepu, Sirkovic, Myaskovsky, Weissenberger, Rong, Tanno, Saab, Popovici, Blum, Zhang, Chou, Hassidim, Gokturk, Vahdat, Kohli, Matias, Carroll, Kulkarni, Tomasev, Guan, Dhillon, Vaishnav, Lee, Costa, Penadés, Peltz, Xu, Pawlosky, Karthikesalingam, and Natarajan]{gottweis2025aicoscientist}
J.~Gottweis, W.-H. Weng, A.~Daryin, T.~Tu, A.~Palepu, P.~Sirkovic, A.~Myaskovsky, F.~Weissenberger, K.~Rong, R.~Tanno, K.~Saab, D.~Popovici, J.~Blum, F.~Zhang, K.~Chou, A.~Hassidim, B.~Gokturk, A.~Vahdat, P.~Kohli, Y.~Matias, A.~Carroll, K.~Kulkarni, N.~Tomasev, Y.~Guan, V.~Dhillon, E.~D. Vaishnav, B.~Lee, T.~R.~D. Costa, J.~R. Penadés, G.~Peltz, Y.~Xu, A.~Pawlosky, A.~Karthikesalingam, and V.~Natarajan.
\newblock Towards an {AI} co-scientist, 2025.
\newblock URL \url{https://arxiv.org/abs/2502.18864}.

\bibitem[Greebel et~al.(2015)Greebel, Moriarty, Callaway, and Xethalis]{greebel2015recent}
E.~L. Greebel, K.~Moriarty, C.~Callaway, and G.~Xethalis.
\newblock Recent key bitcoin and virtual currency regulatory and law enforcement developments.
\newblock \emph{Journal of Investment Compliance}, 16\penalty0 (1):\penalty0 13--18, 2015.

\bibitem[Gu et~al.(2024)Gu, Jiang, Shi, Tan, Zhai, Xu, Li, Shen, Ma, Liu, et~al.]{gu2024survey}
J.~Gu, X.~Jiang, Z.~Shi, H.~Tan, X.~Zhai, C.~Xu, W.~Li, Y.~Shen, S.~Ma, H.~Liu, et~al.
\newblock A survey on llm-as-a-judge.
\newblock \emph{arXiv preprint arXiv:2411.15594}, 2024.

\bibitem[Guo et~al.(2024)Guo, Chen, Wang, Chang, Pei, Chawla, Wiest, and Zhang]{guo2024large}
T.~Guo, X.~Chen, Y.~Wang, R.~Chang, S.~Pei, N.~V. Chawla, O.~Wiest, and X.~Zhang.
\newblock Large language model based multi-agents: A survey of progress and challenges.
\newblock \emph{arXiv preprint arXiv:2402.01680}, 2024.

\bibitem[Gupta et~al.(2023)Gupta, Chandak, Jordan, Thomas, and C~da Silva]{gupta2023behavior}
D.~Gupta, Y.~Chandak, S.~Jordan, P.~S. Thomas, and B.~C~da Silva.
\newblock Behavior alignment via reward function optimization.
\newblock \emph{Advances in Neural Information Processing Systems}, 36:\penalty0 52759--52791, 2023.

\bibitem[Hadfield et~al.(2023)Hadfield, Cu{\'e}llar, and O’Reilly]{hadfield2023s}
G.~Hadfield, M.-F.~T. Cu{\'e}llar, and T.~O’Reilly.
\newblock It's time to create a national registry for large ai models.
\newblock 2023.

\bibitem[Hadfield and Koh(2025)]{hadfield2025economy}
G.~K. Hadfield and A.~Koh.
\newblock An economy of ai agents.
\newblock \emph{arXiv preprint arXiv:2509.01063}, 2025.

\bibitem[Hajiesmaili et~al.(2024)Hajiesmaili, Ren, Sitaraman, and Wierman]{hajiesmaili2024towards}
M.~Hajiesmaili, S.~Ren, R.~K. Sitaraman, and A.~Wierman.
\newblock Towards environmentally equitable ai.
\newblock \emph{arXiv preprint arXiv:2412.16539}, 2024.

\bibitem[Halpern and Shah(2021)]{halpern2021fair}
D.~Halpern and N.~Shah.
\newblock Fair and efficient resource allocation with partial information.
\newblock \emph{arXiv preprint arXiv:2105.10064}, 2021.

\bibitem[Hammond et~al.(2025)Hammond, Chan, Clifton, Hoelscher-Obermaier, Khan, McLean, Smith, Barfuss, Foerster, Gavenčiak, Han, Hughes, Kovařík, Kulveit, Leibo, Oesterheld, de~Witt, Shah, Wellman, Bova, Cimpeanu, Ezell, Feuillade-Montixi, Franklin, Kran, Krawczuk, Lamparth, Lauffer, Meinke, Motwani, Reuel, Conitzer, Dennis, Gabriel, Gleave, Hadfield, Haghtalab, Kasirzadeh, Krier, Larson, Lehman, Parkes, Piliouras, and Rahwan]{hammond2025multiagentrisksadvancedai}
L.~Hammond, A.~Chan, J.~Clifton, J.~Hoelscher-Obermaier, A.~Khan, E.~McLean, C.~Smith, W.~Barfuss, J.~Foerster, T.~Gavenčiak, T.~A. Han, E.~Hughes, V.~Kovařík, J.~Kulveit, J.~Z. Leibo, C.~Oesterheld, C.~S. de~Witt, N.~Shah, M.~Wellman, P.~Bova, T.~Cimpeanu, C.~Ezell, Q.~Feuillade-Montixi, M.~Franklin, E.~Kran, I.~Krawczuk, M.~Lamparth, N.~Lauffer, A.~Meinke, S.~Motwani, A.~Reuel, V.~Conitzer, M.~Dennis, I.~Gabriel, A.~Gleave, G.~Hadfield, N.~Haghtalab, A.~Kasirzadeh, S.~Krier, K.~Larson, J.~Lehman, D.~C. Parkes, G.~Piliouras, and I.~Rahwan.
\newblock Multi-agent risks from advanced ai, 2025.
\newblock URL \url{https://arxiv.org/abs/2502.14143}.

\bibitem[Hao et~al.(2016)Hao, Leung, Hao, and Leung]{hao2016fairness}
J.~Hao, H.-f. Leung, J.~Hao, and H.-f. Leung.
\newblock Fairness in cooperative multiagent systems.
\newblock \emph{Interactions in Multiagent Systems: Fairness, Social Optimality and Individual Rationality}, pages 27--70, 2016.

\bibitem[Hardy and Van~Vugt(2006)]{hardy2006nice}
C.~L. Hardy and M.~Van~Vugt.
\newblock Nice guys finish first: The competitive altruism hypothesis.
\newblock \emph{Personality and Social Psychology Bulletin}, 32\penalty0 (10):\penalty0 1402--1413, 2006.

\bibitem[Harland et~al.(2024)Harland, Dazeley, Vamplew, Senaratne, Nakisa, and Cruz]{harland2024adaptive}
H.~Harland, R.~Dazeley, P.~Vamplew, H.~Senaratne, B.~Nakisa, and F.~Cruz.
\newblock Adaptive alignment: Dynamic preference adjustments via multi-objective reinforcement learning for pluralistic ai.
\newblock \emph{arXiv preprint arXiv:2410.23630}, 2024.

\bibitem[Hayenhjelm(2012)]{hayenhjelm2012fair}
M.~Hayenhjelm.
\newblock What is a fair distribution of risk?
\newblock 2012.

\bibitem[Henrich et~al.(2015)Henrich, Chudek, and Boyd]{henrich2015big}
J.~Henrich, M.~Chudek, and R.~Boyd.
\newblock The big man mechanism: how prestige fosters cooperation and creates prosocial leaders.
\newblock \emph{Philosophical Transactions of the Royal Society B: Biological Sciences}, 370\penalty0 (1683):\penalty0 20150013, 2015.

\bibitem[Hertz et~al.(2025)Hertz, K{\"o}ster, Janssen, and Leibo]{hertz2025beyond}
U.~Hertz, R.~K{\"o}ster, M.~A. Janssen, and J.~Z. Leibo.
\newblock Beyond the matrix: Experimental approaches to studying cognitive agents in social-ecological systems.
\newblock \emph{Cognition}, 254:\penalty0 105993, 2025.

\bibitem[Hettiarachchi(2025)]{hettiarachchi2025exploring}
I.~Hettiarachchi.
\newblock Exploring generative {AI} agents: Architecture, applications, and challenges.
\newblock \emph{Journal of Artificial Intelligence General science (JAIGS) ISSN: 3006-4023}, 8\penalty0 (1):\penalty0 105--127, 2025.

\bibitem[Hosseini et~al.(2025)Hosseini, Khanna, and Singh]{hosseini2025matchingmarketsmeetllms}
H.~Hosseini, S.~Khanna, and R.~Singh.
\newblock Matching markets meet llms: Algorithmic reasoning with ranked preferences, 2025.
\newblock URL \url{https://arxiv.org/abs/2506.04478}.

\bibitem[Hostallero et~al.(2020)Hostallero, Kim, Moon, Son, Kang, and Yi]{hostallero2020inducing}
D.~E. Hostallero, D.~Kim, S.~Moon, K.~Son, W.~J. Kang, and Y.~Yi.
\newblock Inducing cooperation through reward reshaping based on peer evaluations in deep multi-agent reinforcement learning.
\newblock In \emph{Proceedings of the 19th International Conference on Autonomous Agents and MultiAgent Systems}, pages 520--528, 2020.

\bibitem[House(2025)]{americaaiplan}
T.~W. House.
\newblock America's ai action plan, 2025.
\newblock URL \url{https://whitehouse.gov/wp-content/uploads/2025/07/Americas-AI-Action-Plan.pdf}.

\bibitem[{House of Commons Committee of Public Accounts}(2025)]{HoC_PAC_AI_Gov_2025}
{House of Commons Committee of Public Accounts}.
\newblock Use of {AI} in government.
\newblock Report, House of Commons, 2025.
\newblock HC (specify number if known).

\bibitem[Howlett(2022)]{howlett2022avoiding}
M.~Howlett.
\newblock Avoiding a panglossian policy science: The need to deal with the darkside of policy-maker and policy-taker behaviour.
\newblock \emph{Public Integrity}, 24\penalty0 (3):\penalty0 306--318, 2022.

\bibitem[Hsieh et~al.(2018)Hsieh, Vergne, Anderson, Lakhani, and Reitzig]{hsieh2018bitcoin}
Y.-Y. Hsieh, J.-P. Vergne, P.~Anderson, K.~Lakhani, and M.~Reitzig.
\newblock Bitcoin and the rise of decentralized autonomous organizations.
\newblock \emph{Journal of Organization Design}, 7\penalty0 (1):\penalty0 1--16, 2018.

\bibitem[Hu et~al.(2023)Hu, Mu, Yu, Ding, Wu, Shao, Chen, Wang, Qiao, and Luo]{hu2023tree}
M.~Hu, Y.~Mu, X.~Yu, M.~Ding, S.~Wu, W.~Shao, Q.~Chen, B.~Wang, Y.~Qiao, and P.~Luo.
\newblock Tree-planner: Efficient close-loop task planning with large language models.
\newblock \emph{arXiv preprint arXiv:2310.08582}, 2023.

\bibitem[Hua et~al.(2024)Hua, Liu, Li, Amayuelas, Chen, Jiang, Jin, Fan, Sun, Wang, et~al.]{hua2024game}
W.~Hua, O.~Liu, L.~Li, A.~Amayuelas, J.~Chen, L.~Jiang, M.~Jin, L.~Fan, F.~Sun, W.~Wang, et~al.
\newblock Game-theoretic llm: Agent workflow for negotiation games.
\newblock \emph{arXiv preprint arXiv:2411.05990}, 2024.

\bibitem[Huang et~al.(2025)Huang, Yu, Ma, Zhong, Feng, Wang, Chen, Peng, Feng, Qin, et~al.]{huang2025survey}
L.~Huang, W.~Yu, W.~Ma, W.~Zhong, Z.~Feng, H.~Wang, Q.~Chen, W.~Peng, X.~Feng, B.~Qin, et~al.
\newblock A survey on hallucination in large language models: Principles, taxonomy, challenges, and open questions.
\newblock \emph{ACM Transactions on Information Systems}, 43\penalty0 (2):\penalty0 1--55, 2025.

\bibitem[Huang et~al.(2024)Huang, Liu, Chen, Wang, Wang, Lian, Wang, Tang, and Chen]{huang2024understanding}
X.~Huang, W.~Liu, X.~Chen, X.~Wang, H.~Wang, D.~Lian, Y.~Wang, R.~Tang, and E.~Chen.
\newblock Understanding the planning of {LLM} agents: A survey.
\newblock \emph{arXiv preprint arXiv:2402.02716}, 2024.

\bibitem[Hughes et~al.(2025)Hughes, Zhu, Chadwick, Koster, Casta{\~n}eda, Beattie, Graepel, Botvinick, and Leibo]{hughes2025modeling}
E.~Hughes, T.~O. Zhu, M.~J. Chadwick, R.~Koster, A.~G. Casta{\~n}eda, C.~Beattie, T.~Graepel, M.~M. Botvinick, and J.~Z. Leibo.
\newblock Modeling human reputation-seeking behavior in a spatio-temporally complex public good provision game.
\newblock \emph{arXiv preprint arXiv:2506.06032}, 2025.

\bibitem[Hughes(2017)]{hughes2017cryptocurrency}
S.~D. Hughes.
\newblock Cryptocurrency regulations and enforcement in the us.
\newblock \emph{W. St. UL Rev.}, 45:\penalty0 1, 2017.

\bibitem[IMDA(2025)]{singlaunch}
IMDA.
\newblock Singapore launches new tools to help businesses protect data and deploy ai in a trusted ecosystem, 2025.
\newblock URL \url{https://www.imda.gov.sg/resources/press-releases-factsheets-and-speeches/press-releases/2025/singapore-launches-new-tools-to-help-businesses-protect-data-and-deploy-ai-in-a-trusted-ecosystem}.

\bibitem[Iyer and Huhns(2005)]{iyer2005multiagent}
K.~Iyer and M.~Huhns.
\newblock Multiagent negotiation for fair and unbiased resource allocation.
\newblock In \emph{OTM Confederated International Conferences" On the Move to Meaningful Internet Systems"}, pages 453--465. Springer, 2005.

\bibitem[Jacobs and Wallach(2021)]{jacobs2021measurement}
A.~Z. Jacobs and H.~Wallach.
\newblock Measurement and fairness.
\newblock In \emph{Proceedings of the 2021 ACM conference on fairness, accountability, and transparency}, pages 375--385, 2021.

\bibitem[Jacobs(2025)]{jacobs2025brookings}
J.~Jacobs.
\newblock Ai labor displacement and the limits of worker retraining.
\newblock 2025.

\bibitem[Jaderberg et~al.(2019)Jaderberg, Czarnecki, Dunning, Marris, Lever, Castaneda, Beattie, Rabinowitz, Morcos, Ruderman, et~al.]{jaderberg2019human}
M.~Jaderberg, W.~M. Czarnecki, I.~Dunning, L.~Marris, G.~Lever, A.~G. Castaneda, C.~Beattie, N.~C. Rabinowitz, A.~S. Morcos, A.~Ruderman, et~al.
\newblock Human-level performance in 3d multiplayer games with population-based reinforcement learning.
\newblock \emph{Science}, 364\penalty0 (6443):\penalty0 859--865, 2019.

\bibitem[Jannelli et~al.(2024)Jannelli, Schoepf, Bickel, Netland, and Brintrup]{jannelli2024agentic}
V.~Jannelli, S.~Schoepf, M.~Bickel, T.~Netland, and A.~Brintrup.
\newblock Agentic llms in the supply chain: Towards autonomous multi-agent consensus-seeking.
\newblock \emph{arXiv preprint arXiv:2411.10184}, 2024.

\bibitem[Jeyasheela~Rakkini and Geetha(2021)]{jeyasheela2021blockchain}
M.~Jeyasheela~Rakkini and K.~Geetha.
\newblock Blockchain-enabled microfinance model with decentralized autonomous organizations.
\newblock In \emph{Computer Networks and Inventive Communication Technologies: Proceedings of Third ICCNCT 2020}, pages 417--430. Springer, 2021.

\bibitem[Ji et~al.(2023)Ji, Qiu, Chen, Zhang, Lou, Wang, Duan, He, Zhou, Zhang, et~al.]{ji2023ai}
J.~Ji, T.~Qiu, B.~Chen, B.~Zhang, H.~Lou, K.~Wang, Y.~Duan, Z.~He, J.~Zhou, Z.~Zhang, et~al.
\newblock Ai alignment: A comprehensive survey.
\newblock \emph{arXiv preprint arXiv:2310.19852}, 2023.

\bibitem[Ji et~al.(2024)Ji, Li, Liu, Du, Wei, Shen, Qi, and Lin]{ji2024srapagentsimulatingoptimizingscarce}
J.~Ji, Y.~Li, H.~Liu, Z.~Du, Z.~Wei, W.~Shen, Q.~Qi, and Y.~Lin.
\newblock Srap-agent: Simulating and optimizing scarce resource allocation policy with llm-based agent, 2024.
\newblock URL \url{https://arxiv.org/abs/2410.14152}.

\bibitem[Jiang and Lu(2019)]{jiang2019learning}
J.~Jiang and Z.~Lu.
\newblock Learning fairness in multi-agent systems.
\newblock \emph{Advances in Neural Information Processing Systems}, 32, 2019.

\bibitem[Jiang et~al.(2024)Jiang, Li, Zhou, Qi, Hu, Wei, Jiang, and Wu]{jiang2024ai}
Y.-H. Jiang, R.~Li, Y.~Zhou, C.~Qi, H.~Hu, Y.~Wei, B.~Jiang, and Y.~Wu.
\newblock Ai agent for education: von neumann multi-agent system framework.
\newblock \emph{arXiv preprint arXiv:2501.00083}, 2024.

\bibitem[Jin et~al.(2025)Jin, Zhang, Wang, and Cong]{jin2025stellaselfevolvingllmagent}
R.~Jin, Z.~Zhang, M.~Wang, and L.~Cong.
\newblock Stella: Self-evolving {LLM} agent for biomedical research, 2025.
\newblock URL \url{https://arxiv.org/abs/2507.02004}.

\bibitem[Johanson et~al.(2022)Johanson, Hughes, Timbers, and Leibo]{johanson2022emergent}
M.~B. Johanson, E.~Hughes, F.~Timbers, and J.~Z. Leibo.
\newblock Emergent bartering behaviour in multi-agent reinforcement learning.
\newblock \emph{arXiv preprint arXiv:2205.06760}, 2022.

\bibitem[Kash et~al.(2014)Kash, Procaccia, and Shah]{kash2014no}
I.~Kash, A.~D. Procaccia, and N.~Shah.
\newblock No agent left behind: Dynamic fair division of multiple resources.
\newblock \emph{Journal of Artificial Intelligence Research}, 51:\penalty0 579--603, 2014.

\bibitem[Kasirzadeh and Gabriel(2025)]{kasirzadeh2025characterizingaiagentsalignment}
A.~Kasirzadeh and I.~Gabriel.
\newblock Characterizing {AI} agents for alignment and governance, 2025.
\newblock URL \url{https://arxiv.org/abs/2504.21848}.

\bibitem[Khanov et~al.(2024)Khanov, Burapacheep, and Li]{khanov2024args}
M.~Khanov, J.~Burapacheep, and Y.~Li.
\newblock Args: Alignment as reward-guided search.
\newblock \emph{arXiv preprint arXiv:2402.01694}, 2024.

\bibitem[Kharrazi et~al.(2020)Kharrazi, Yu, Jacob, Vora, and Fath]{kharrazi2020redundancy}
A.~Kharrazi, Y.~Yu, A.~Jacob, N.~Vora, and B.~D. Fath.
\newblock Redundancy, diversity, and modularity in network resilience: applications for international trade and implications for public policy.
\newblock \emph{Current research in environmental sustainability}, 2:\penalty0 100006, 2020.

\bibitem[Kichiji and Nishibe(2008)]{kichiji2008network}
N.~Kichiji and M.~Nishibe.
\newblock Network analyses of the circulation flow of community currency.
\newblock \emph{Evolutionary and Institutional Economics Review}, 4:\penalty0 267--300, 2008.

\bibitem[Kirchherr et~al.(2023)Kirchherr, Hartley, and Tukker]{kirchherr2023missions}
J.~Kirchherr, K.~Hartley, and A.~Tukker.
\newblock Missions and mission-oriented innovation policy for sustainability: A review and critical reflection.
\newblock \emph{Environmental Innovation and Societal Transitions}, 47:\penalty0 100721, 2023.

\bibitem[Kirilenko et~al.(2017)Kirilenko, Kyle, Samadi, and Tuzun]{kirilenko2017flash}
A.~Kirilenko, A.~S. Kyle, M.~Samadi, and T.~Tuzun.
\newblock The flash crash: High-frequency trading in an electronic market.
\newblock \emph{The Journal of Finance}, 72\penalty0 (3):\penalty0 967--998, 2017.

\bibitem[Kirk et~al.(2023)Kirk, Vidgen, R{\"o}ttger, and Hale]{kirk2023personalisation}
H.~R. Kirk, B.~Vidgen, P.~R{\"o}ttger, and S.~A. Hale.
\newblock Personalisation within bounds: A risk taxonomy and policy framework for the alignment of large language models with personalised feedback.
\newblock \emph{arXiv preprint arXiv:2303.05453}, 2023.

\bibitem[Kirk et~al.(2024)Kirk, Vidgen, R{\"o}ttger, and Hale]{kirk2024benefits}
H.~R. Kirk, B.~Vidgen, P.~R{\"o}ttger, and S.~A. Hale.
\newblock The benefits, risks and bounds of personalizing the alignment of large language models to individuals.
\newblock \emph{Nature Machine Intelligence}, 6\penalty0 (4):\penalty0 383--392, 2024.

\bibitem[Kohn(2003)]{kohn2003financial}
M.~Kohn.
\newblock Financial institutions and markets.
\newblock \emph{OUP Catalogue}, 2003.

\bibitem[Koster et~al.(2022)Koster, Balaguer, Tacchetti, Weinstein, Zhu, Hauser, Williams, Campbell-Gillingham, Thacker, Botvinick, et~al.]{koster2022human}
R.~Koster, J.~Balaguer, A.~Tacchetti, A.~Weinstein, T.~Zhu, O.~Hauser, D.~Williams, L.~Campbell-Gillingham, P.~Thacker, M.~Botvinick, et~al.
\newblock Human-centred mechanism design with democratic ai.
\newblock \emph{Nature Human Behaviour}, 6\penalty0 (10):\penalty0 1398--1407, 2022.

\bibitem[K{\"o}ster et~al.(2025)K{\"o}ster, Du{\'e}{\~n}ez-Guzm{\'a}n, Cunningham, and Leibo]{koster2025tabula}
R.~K{\"o}ster, E.~A. Du{\'e}{\~n}ez-Guzm{\'a}n, W.~A. Cunningham, and J.~Z. Leibo.
\newblock Tabula rasa agents display emergent in-group behavior.
\newblock \emph{Proceedings of the National Academy of Sciences}, 122\penalty0 (25):\penalty0 e2319947121, 2025.

\bibitem[Kulveit et~al.(2025)Kulveit, Douglas, Ammann, Turan, Krueger, and Duvenaud]{kulveit2025gradual}
J.~Kulveit, R.~Douglas, N.~Ammann, D.~Turan, D.~Krueger, and D.~Duvenaud.
\newblock Gradual disempowerment: Systemic existential risks from incremental {AI} development.
\newblock \emph{arXiv preprint arXiv:2501.16946}, 2025.

\bibitem[Kumar and Yeoh(2025)]{kumar2025decaf}
A.~Kumar and W.~Yeoh.
\newblock Decaf: Learning to be fair in multi-agent resource allocation.
\newblock \emph{arXiv preprint arXiv:2502.04281}, 2025.

\bibitem[Kyriakou and Otterbacher(2023)]{kyriakou2023humans}
K.~Kyriakou and J.~Otterbacher.
\newblock In humans, we trust: Multidisciplinary perspectives on the requirements for human oversight in algorithmic processes.
\newblock \emph{Discover Artificial Intelligence}, 3\penalty0 (1):\penalty0 44, 2023.

\bibitem[Lee and L’heureux(2020)]{lee2020regulatory}
J.~Lee and F.~L’heureux.
\newblock A regulatory framework for cryptocurrency.
\newblock \emph{European Business Law Review}, 31\penalty0 (3), 2020.

\bibitem[Lee(2009)]{lee2009fairness}
S.~Lee.
\newblock Fairness, stability and optimality of adaptive multiagent systems: Interaction through resource sharing.
\newblock \emph{IEEE transactions on automation science and engineering}, 7\penalty0 (3):\penalty0 427--439, 2009.

\bibitem[Lehman(2023)]{lehman2023machinelove}
J.~Lehman.
\newblock Machine love, 2023.
\newblock URL \url{https://arxiv.org/abs/2302.09248}.

\bibitem[Leibo et~al.(2019)Leibo, Hughes, Lanctot, and Graepel]{leibo2019autocurricula}
J.~Z. Leibo, E.~Hughes, M.~Lanctot, and T.~Graepel.
\newblock Autocurricula and the emergence of innovation from social interaction: A manifesto for multi-agent intelligence research.
\newblock \emph{arXiv preprint arXiv:1903.00742}, 2019.

\bibitem[Leibo et~al.(2024)Leibo, Vezhnevets, Diaz, Agapiou, Cunningham, Sunehag, Haas, Koster, Du{\'e}{\~n}ez-Guzm{\'a}n, Isaac, Piliouras, Bileschi, Rahwan, and Osindero]{leibo2024theory}
J.~Z. Leibo, A.~S. Vezhnevets, M.~Diaz, J.~P. Agapiou, W.~A. Cunningham, P.~Sunehag, J.~Haas, R.~Koster, E.~A. Du{\'e}{\~n}ez-Guzm{\'a}n, W.~S. Isaac, G.~Piliouras, S.~M. Bileschi, I.~Rahwan, and S.~Osindero.
\newblock A theory of appropriateness with applications to generative artificial intelligence.
\newblock \emph{arXiv preprint arXiv:2412.19010}, 2024.
\newblock URL \url{https://doi.org/10.48550/arXiv.2412.19010}.

\bibitem[Leibo et~al.(2025)Leibo, Vezhnevets, Cunningham, Krier, Diaz, and Osindero]{leibo2025societal}
J.~Z. Leibo, A.~S. Vezhnevets, W.~A. Cunningham, S.~Krier, M.~Diaz, and S.~Osindero.
\newblock Societal and technological progress as sewing an ever-growing, ever-changing, patchy, and polychrome quilt.
\newblock \emph{arXiv preprint arXiv:2505.05197}, 2025.

\bibitem[Leike et~al.(2018)Leike, Krueger, Everitt, Martic, Maini, and Legg]{leike2018scalable}
J.~Leike, D.~Krueger, T.~Everitt, M.~Martic, V.~Maini, and S.~Legg.
\newblock Scalable agent alignment via reward modeling: a research direction.
\newblock \emph{arXiv preprint arXiv:1811.07871}, 2018.

\bibitem[Li and Lee(2022)]{li2022modeling}
M.~Li and J.~D. Lee.
\newblock Modeling goal alignment in human-ai teaming: a dynamic game theory approach.
\newblock In \emph{Proceedings of the Human Factors and Ergonomics Society Annual Meeting}, volume~66, pages 1538--1542. SAGE Publications Sage CA: Los Angeles, CA, 2022.

\bibitem[Li et~al.(2024{\natexlab{a}})Li, Yang, Wierman, and Ren]{li2024towards}
P.~Li, J.~Yang, A.~Wierman, and S.~Ren.
\newblock Towards environmentally equitable {AI} via geographical load balancing.
\newblock In \emph{Proceedings of the 15th ACM International Conference on Future and Sustainable Energy Systems}, pages 291--307, 2024{\natexlab{a}}.

\bibitem[Li et~al.(2024{\natexlab{b}})Li, Sun, Cheng, and Qiu]{li2024agent}
S.~Li, T.~Sun, Q.~Cheng, and X.~Qiu.
\newblock Agent alignment in evolving social norms.
\newblock \emph{arXiv preprint arXiv:2401.04620}, 2024{\natexlab{b}}.

\bibitem[Li et~al.(2024{\natexlab{c}})Li, Zhang, Sun, Zhang, Du, Wen, Wang, and Pan]{li2024tackling}
Y.~Li, S.~Zhang, J.~Sun, W.~Zhang, Y.~Du, Y.~Wen, X.~Wang, and W.~Pan.
\newblock Tackling cooperative incompatibility for zero-shot human-ai coordination.
\newblock \emph{Journal of Artificial Intelligence Research}, 80:\penalty0 1139--1185, 2024{\natexlab{c}}.

\bibitem[Liang et~al.(2020)Liang, Kamat, and Menassa]{liang2020teaching}
C.-J. Liang, V.~R. Kamat, and C.~C. Menassa.
\newblock Teaching robots to perform quasi-repetitive construction tasks through human demonstration.
\newblock \emph{Automation in Construction}, 120:\penalty0 103370, 2020.

\bibitem[List(2021)]{list2021group}
C.~List.
\newblock Group agency and artificial intelligence.
\newblock \emph{Philosophy \& technology}, 34\penalty0 (4):\penalty0 1213--1242, 2021.

\bibitem[List and Pettit(2011)]{list2011group}
C.~List and P.~Pettit.
\newblock \emph{Group agency: The possibility, design, and status of corporate agents}.
\newblock Oxford University Press, 2011.

\bibitem[Liu et~al.(2023{\natexlab{a}})Liu, Jiang, Zhang, Liu, Zhang, Biswas, and Stone]{liu2023llm+}
B.~Liu, Y.~Jiang, X.~Zhang, Q.~Liu, S.~Zhang, J.~Biswas, and P.~Stone.
\newblock Llm+ p: Empowering large language models with optimal planning proficiency.
\newblock \emph{arXiv preprint arXiv:2304.11477}, 2023{\natexlab{a}}.

\bibitem[Liu et~al.(2024{\natexlab{a}})Liu, Wang, Chen, Peng, Chen, Zhang, and Lou]{liu2024large}
J.~Liu, K.~Wang, Y.~Chen, X.~Peng, Z.~Chen, L.~Zhang, and Y.~Lou.
\newblock Large language model-based agents for software engineering: A survey.
\newblock \emph{arXiv preprint arXiv:2409.02977}, 2024{\natexlab{a}}.

\bibitem[Liu et~al.(2024{\natexlab{b}})Liu, Xiong, Ren, Chen, Wu, Joshi, Gao, Shen, Qin, Yu, et~al.]{liu2024rrm}
T.~Liu, W.~Xiong, J.~Ren, L.~Chen, J.~Wu, R.~Joshi, Y.~Gao, J.~Shen, Z.~Qin, T.~Yu, et~al.
\newblock Rrm: Robust reward model training mitigates reward hacking.
\newblock \emph{arXiv preprint arXiv:2409.13156}, 2024{\natexlab{b}}.

\bibitem[Liu et~al.(2023{\natexlab{b}})Liu, Yu, Zhang, Xu, Lei, Lai, Gu, Ding, Men, Yang, Zhang, Deng, Zeng, Du, Zhang, Shen, Zhang, Su, Sun, Huang, Dong, and Tang]{liu2023agentbenchevaluatingllmsagents}
X.~Liu, H.~Yu, H.~Zhang, Y.~Xu, X.~Lei, H.~Lai, Y.~Gu, H.~Ding, K.~Men, K.~Yang, S.~Zhang, X.~Deng, A.~Zeng, Z.~Du, C.~Zhang, S.~Shen, T.~Zhang, Y.~Su, H.~Sun, M.~Huang, Y.~Dong, and J.~Tang.
\newblock Agentbench: Evaluating llms as agents, 2023{\natexlab{b}}.
\newblock URL \url{https://arxiv.org/abs/2308.03688}.

\bibitem[Liu et~al.(2024{\natexlab{c}})Liu, Yi, Chen, Yao, Yi, Zan, Liu, Xie, and Ho]{liu2024elephant}
Y.~Liu, X.~Yi, X.~Chen, J.~Yao, J.~Yi, D.~Zan, Z.~Liu, X.~Xie, and T.-Y. Ho.
\newblock Elephant in the room: Unveiling the impact of reward model quality in alignment.
\newblock \emph{arXiv preprint arXiv:2409.19024}, 2024{\natexlab{c}}.

\bibitem[Luccioni and Hernandez-Garcia(2023)]{luccioni2023counting}
A.~S. Luccioni and A.~Hernandez-Garcia.
\newblock Counting carbon: A survey of factors influencing the emissions of machine learning.
\newblock \emph{arXiv preprint arXiv:2302.08476}, 2023.

\bibitem[Luccioni et~al.(2025)Luccioni, Strubell, and Crawford]{luccioni2025efficiency}
A.~S. Luccioni, E.~Strubell, and K.~Crawford.
\newblock From efficiency gains to rebound effects: The problem of jevons' paradox in ai's polarized environmental debate.
\newblock \emph{arXiv preprint arXiv:2501.16548}, 2025.

\bibitem[Luccioni et~al.(2024)Luccioni, Jernite, and Strubell]{luccioni2024power}
S.~Luccioni, Y.~Jernite, and E.~Strubell.
\newblock Power hungry processing: Watts driving the cost of {AI} deployment?
\newblock In \emph{Proceedings of the 2024 ACM conference on fairness, accountability, and transparency}, pages 85--99, 2024.

\bibitem[Lujak et~al.(2023)Lujak, Fern{\'a}ndez, Billhardt, Ossowski, Arias, and L{\'o}pez~S{\'a}nchez]{lujak2023value}
M.~Lujak, A.~Fern{\'a}ndez, H.~Billhardt, S.~Ossowski, J.~Arias, and A.~L{\'o}pez~S{\'a}nchez.
\newblock On value-aligned cooperative multi-agent task allocation.
\newblock In \emph{International Workshop on Value Engineering in AI}, pages 197--216. Springer, 2023.

\bibitem[Lythreatis et~al.(2022)Lythreatis, Singh, and El-Kassar]{lythreatis2022digital}
S.~Lythreatis, S.~K. Singh, and A.-N. El-Kassar.
\newblock The digital divide: A review and future research agenda.
\newblock \emph{Technological Forecasting and Social Change}, 175:\penalty0 121359, 2022.

\bibitem[Magni and Milella(2025)]{magni2025conversational}
M.~Magni and F.~Milella.
\newblock Conversational agents in the legal domain: A systematic review.
\newblock In \emph{Advances in Information and Communication: Proceedings of the 2025 Future of Information and Communication Conference (FICC), Volume 3}, volume 1285, page 183. Springer Nature, 2025.

\bibitem[Marcoux and Lusseau(2013)]{marcoux2013network}
M.~Marcoux and D.~Lusseau.
\newblock Network modularity promotes cooperation.
\newblock \emph{Journal of Theoretical Biology}, 324:\penalty0 103--108, 2013.

\bibitem[Mattsson et~al.(2023)Mattsson, Criscione, and Takes]{mattsson2023circulation}
C.~E. Mattsson, T.~Criscione, and F.~W. Takes.
\newblock Circulation of a digital community currency.
\newblock \emph{Scientific Reports}, 13\penalty0 (1):\penalty0 5864, 2023.

\bibitem[Mazzocca et~al.(2025)Mazzocca, Acar, Uluagac, Montanari, Bellavista, and Conti]{mazzocca2025survey}
C.~Mazzocca, A.~Acar, S.~Uluagac, R.~Montanari, P.~Bellavista, and M.~Conti.
\newblock A survey on decentralized identifiers and verifiable credentials.
\newblock \emph{IEEE Communications Surveys \& Tutorials}, 2025.

\bibitem[Mazzucato(2015)]{mazzucato2015building}
M.~Mazzucato.
\newblock Building the entrepreneurial state: A new framework for envisioning and evaluating a mission-oriented public sector.
\newblock \emph{Levy Economics Institute of Bard College Working Paper}, \penalty0 (824), 2015.

\bibitem[Mazzucato(2018)]{mazzucato2018mission}
M.~Mazzucato.
\newblock Mission-oriented innovation policies: challenges and opportunities.
\newblock \emph{Industrial and corporate change}, 27\penalty0 (5):\penalty0 803--815, 2018.

\bibitem[Mazzucato(2023)]{mazzucato2023financing}
M.~Mazzucato.
\newblock Financing the sustainable development goals through mission-oriented development banks.
\newblock 2023.

\bibitem[McCormick(2010)]{mccormick2010legal}
R.~McCormick.
\newblock \emph{Legal risk in the financial markets}.
\newblock Oxford University Press, USA, 2010.

\bibitem[Menkveld and Yueshen(2019)]{menkveld2019flash}
A.~J. Menkveld and B.~Z. Yueshen.
\newblock The flash crash: A cautionary tale about highly fragmented markets.
\newblock \emph{Management Science}, 65\penalty0 (10):\penalty0 4470--4488, 2019.

\bibitem[Metcalf(2009)]{metcalf2009market}
G.~E. Metcalf.
\newblock Market-based policy options to control us greenhouse gas emissions.
\newblock \emph{Journal of Economic perspectives}, 23\penalty0 (2):\penalty0 5--27, 2009.

\bibitem[Michel and Hudon(2015)]{michel2015community}
A.~Michel and M.~Hudon.
\newblock Community currencies and sustainable development: A systematic review.
\newblock \emph{Ecological economics}, 116:\penalty0 160--171, 2015.

\bibitem[Microsoft(2025)]{mcp2}
Microsoft.
\newblock Unleashing the power of model context protocol (mcp): A game-changer in {AI} integration, 2025.

\bibitem[Mishra(2023)]{mishra2023ai}
A.~Mishra.
\newblock Ai alignment and social choice: Fundamental limitations and policy implications.
\newblock \emph{arXiv preprint arXiv:2310.16048}, 2023.

\bibitem[Mokyr et~al.(2015)Mokyr, Vickers, and Ziebarth]{mokyr2015history}
J.~Mokyr, C.~Vickers, and N.~L. Ziebarth.
\newblock The history of technological anxiety and the future of economic growth: Is this time different?
\newblock \emph{Journal of economic perspectives}, 29\penalty0 (3):\penalty0 31--50, 2015.

\bibitem[Moloney(2023)]{moloney2023eu}
N.~Moloney.
\newblock \emph{EU securities and financial markets regulation}.
\newblock Oxford University Press, 2023.

\bibitem[Mu et~al.(2024)Mu, Guo, Chen, Shen, Hu, Hu, and Wang]{mu2024multi}
C.~Mu, H.~Guo, Y.~Chen, C.~Shen, D.~Hu, S.~Hu, and Z.~Wang.
\newblock Multi-agent, human--agent and beyond: a survey on cooperation in social dilemmas.
\newblock \emph{Neurocomputing}, 610:\penalty0 128514, 2024.

\bibitem[Ngo et~al.(2022)Ngo, Chan, and Mindermann]{ngo2022alignment}
R.~Ngo, L.~Chan, and S.~Mindermann.
\newblock The alignment problem from a deep learning perspective.
\newblock \emph{arXiv preprint arXiv:2209.00626}, 2022.

\bibitem[Norheim(2016)]{norheim2016ethical}
O.~F. Norheim.
\newblock Ethical priority setting for universal health coverage: challenges in deciding upon fair distribution of health services.
\newblock \emph{BMC medicine}, 14:\penalty0 1--4, 2016.

\bibitem[Ostrom(1990)]{ostrom1990governing}
E.~Ostrom.
\newblock \emph{Governing the commons: The evolution of institutions for collective action}.
\newblock Cambridge university press, 1990.

\bibitem[Paccagnan et~al.(2022)Paccagnan, Chandan, and Marden]{paccagnan2022utility}
D.~Paccagnan, R.~Chandan, and J.~R. Marden.
\newblock Utility and mechanism design in multi-agent systems: An overview.
\newblock \emph{Annual Reviews in Control}, 53:\penalty0 315--328, 2022.

\bibitem[Page(2011)]{page2011climatic}
E.~A. Page.
\newblock Climatic justice and the fair distribution of atmospheric burdens: A conjunctive account.
\newblock \emph{The monist}, 94\penalty0 (3):\penalty0 412--432, 2011.

\bibitem[Pan et~al.(2022)Pan, Bhatia, and Steinhardt]{pan2022effects}
A.~Pan, K.~Bhatia, and J.~Steinhardt.
\newblock The effects of reward misspecification: Mapping and mitigating misaligned models.
\newblock \emph{arXiv preprint arXiv:2201.03544}, 2022.

\bibitem[Pan et~al.(2024)Pan, Gao, Xie, Chen, Wei, Li, Ding, Wen, and Zhou]{pan2024largescalemultiagentsimulationagentscope}
X.~Pan, D.~Gao, Y.~Xie, Y.~Chen, Z.~Wei, Y.~Li, B.~Ding, J.-R. Wen, and J.~Zhou.
\newblock Very large-scale multi-agent simulation in agentscope, 2024.
\newblock URL \url{https://arxiv.org/abs/2407.17789}.

\bibitem[Papoudakis et~al.(2019)Papoudakis, Christianos, Rahman, and Albrecht]{papoudakis2019dealing}
G.~Papoudakis, F.~Christianos, A.~Rahman, and S.~V. Albrecht.
\newblock Dealing with non-stationarity in multi-agent deep reinforcement learning.
\newblock \emph{arXiv preprint arXiv:1906.04737}, 2019.

\bibitem[Park et~al.(2023)Park, Leahey, and Funk]{park2023papers}
M.~Park, E.~Leahey, and R.~J. Funk.
\newblock Papers and patents are becoming less disruptive over time.
\newblock \emph{Nature}, 613\penalty0 (7942):\penalty0 138--144, 2023.

\bibitem[Patel et~al.(2025)Patel, Raut, Cheetirala, Glicksberg, Levin, Nadkarni, Freeman, Klang, and Timsina]{patel2025ai}
D.~Patel, G.~Raut, S.~N. Cheetirala, B.~Glicksberg, M.~A. Levin, G.~Nadkarni, R.~Freeman, E.~Klang, and P.~Timsina.
\newblock Ai agents in modern healthcare: From foundation to pioneer--a comprehensive review and implementation roadmap for impact and integration in clinical settings.
\newblock 2025.

\bibitem[Pedersen et~al.(2003)Pedersen, Kortenkamp, Wettergreen, and Nourbakhsh]{pedersen2003survey}
L.~Pedersen, D.~Kortenkamp, D.~Wettergreen, and I.~Nourbakhsh.
\newblock A survey of space robotics.
\newblock In \emph{Proceeding of the 7th International Symposium on Artificial Intelligence, Robotics and Automation in Space}, number AM-11. European Space Agency, 2003.

\bibitem[Perolat et~al.(2017)Perolat, Leibo, Zambaldi, Beattie, Tuyls, and Graepel]{perolat2017multi}
J.~Perolat, J.~Z. Leibo, V.~Zambaldi, C.~Beattie, K.~Tuyls, and T.~Graepel.
\newblock A multi-agent reinforcement learning model of common-pool resource appropriation.
\newblock \emph{Advances in neural information processing systems}, 30, 2017.

\bibitem[Piatti et~al.(2024)Piatti, Jin, Kleiman-Weiner, Sch{\"o}lkopf, Sachan, and Mihalcea]{piatti2024cooperate}
G.~Piatti, Z.~Jin, M.~Kleiman-Weiner, B.~Sch{\"o}lkopf, M.~Sachan, and R.~Mihalcea.
\newblock Cooperate or collapse: Emergence of sustainable cooperation in a society of {LLM} agents.
\newblock \emph{Advances in Neural Information Processing Systems}, 37:\penalty0 111715--111759, 2024.

\bibitem[Qian et~al.(2024)Qian, Xie, Wang, Liu, Dang, Du, Chen, Yang, Liu, and Sun]{qian2024scaling}
C.~Qian, Z.~Xie, Y.~Wang, W.~Liu, Y.~Dang, Z.~Du, W.~Chen, C.~Yang, Z.~Liu, and M.~Sun.
\newblock Scaling large-language-model-based multi-agent collaboration.
\newblock \emph{arXiv preprint arXiv:2406.07155}, 2024.

\bibitem[Raab(2024)]{raab2024machine}
R.~Raab.
\newblock \emph{Machine Learning and the Multiagent Alignment Problem}.
\newblock University of California, Santa Cruz, 2024.

\bibitem[R{\u{a}}dulescu et~al.(2020)R{\u{a}}dulescu, Mannion, Roijers, and Now{\'e}]{ruadulescu2020multi}
R.~R{\u{a}}dulescu, P.~Mannion, D.~M. Roijers, and A.~Now{\'e}.
\newblock Multi-objective multi-agent decision making: a utility-based analysis and survey.
\newblock \emph{Autonomous Agents and Multi-Agent Systems}, 34\penalty0 (1):\penalty0 10, 2020.

\bibitem[Raileanu et~al.(2018)Raileanu, Denton, Szlam, and Fergus]{raileanu2018modeling}
R.~Raileanu, E.~Denton, A.~Szlam, and R.~Fergus.
\newblock Modeling others using oneself in multi-agent reinforcement learning.
\newblock In \emph{International conference on machine learning}, pages 4257--4266. PMLR, 2018.

\bibitem[Ramirez(2025)]{gptwords}
V.~B. Ramirez.
\newblock Chatgpt is changing the words we use in conversation, 2025.
\newblock URL \url{https://www.scientificamerican.com/article/chatgpt-is-changing-the-words-we-use-in-conversation/}.

\bibitem[Rasal and Hauer(2024)]{rasal2024navigating}
S.~Rasal and E.~Hauer.
\newblock Navigating complexity: Orchestrated problem solving with multi-agent llms.
\newblock \emph{arXiv preprint arXiv:2402.16713}, 2024.

\bibitem[Ren et~al.(2025)Ren, Fu, Zou, Shen, Cai, Chu, Wang, and Hu]{ren2025tragedycommonsbuildingreputation}
S.~Ren, W.~Fu, X.~Zou, C.~Shen, Y.~Cai, C.~Chu, Z.~Wang, and S.~Hu.
\newblock Beyond the tragedy of the commons: Building a reputation system for generative multi-agent systems, 2025.
\newblock URL \url{https://arxiv.org/abs/2505.05029}.

\bibitem[Ren et~al.(2005)Ren, Beard, and Atkins]{ren2005survey}
W.~Ren, R.~W. Beard, and E.~M. Atkins.
\newblock A survey of consensus problems in multi-agent coordination.
\newblock In \emph{Proceedings of the 2005, American Control Conference, 2005.}, pages 1859--1864. IEEE, 2005.

\bibitem[Ruan et~al.(2023)Ruan, Chen, Zhang, Xu, Bao, Mao, Li, Zeng, Zhao, et~al.]{ruan2023tptu}
J.~Ruan, Y.~Chen, B.~Zhang, Z.~Xu, T.~Bao, H.~Mao, Z.~Li, X.~Zeng, R.~Zhao, et~al.
\newblock Tptu: Task planning and tool usage of large language model-based {AI} agents.
\newblock In \emph{NeurIPS 2023 Foundation Models for Decision Making Workshop}, 2023.

\bibitem[Saab et~al.(2025)Saab, Freyberg, Park, Strother, Cheng, Weng, Barrett, Stutz, Tomasev, Palepu, Liévin, Sharma, Ruparel, Ahmed, Vedadi, Kanada, Hughes, Liu, Brown, Gao, Li, Mahdavi, Manyika, Chou, Matias, Hassidim, Webster, Kohli, Eslami, Barral, Rodman, Natarajan, Schaekermann, Tu, Karthikesalingam, and Tanno]{saab2025advancingconversationaldiagnosticai}
K.~Saab, J.~Freyberg, C.~Park, T.~Strother, Y.~Cheng, W.-H. Weng, D.~G.~T. Barrett, D.~Stutz, N.~Tomasev, A.~Palepu, V.~Liévin, Y.~Sharma, R.~Ruparel, A.~Ahmed, E.~Vedadi, K.~Kanada, C.~Hughes, Y.~Liu, G.~Brown, Y.~Gao, S.~Li, S.~S. Mahdavi, J.~Manyika, K.~Chou, Y.~Matias, A.~Hassidim, D.~R. Webster, P.~Kohli, S.~M.~A. Eslami, J.~Barral, A.~Rodman, V.~Natarajan, M.~Schaekermann, T.~Tu, A.~Karthikesalingam, and R.~Tanno.
\newblock Advancing conversational diagnostic {AI} with multimodal reasoning, 2025.
\newblock URL \url{https://arxiv.org/abs/2505.04653}.

\bibitem[Sager et~al.(2025)Sager, Meyer, Yan, von Wartburg-Kottler, Etaiwi, Enayati, Nobel, Abdulkadir, Grewe, and Stadelmann]{sager2025ai}
P.~J. Sager, B.~Meyer, P.~Yan, R.~von Wartburg-Kottler, L.~Etaiwi, A.~Enayati, G.~Nobel, A.~Abdulkadir, B.~F. Grewe, and T.~Stadelmann.
\newblock Ai agents for computer use: A review of instruction-based computer control, gui automation, and operator assistants.
\newblock \emph{arXiv preprint arXiv:2501.16150}, 2025.

\bibitem[Sanabria and Vecino(2025)]{sanabria2025sumunlockingaiagents}
J.~M. Sanabria and P.~A. Vecino.
\newblock Beyond the sum: Unlocking {AI} agents potential through market forces, 2025.
\newblock URL \url{https://arxiv.org/abs/2501.10388}.

\bibitem[Sandel(1998)]{sandel1998money}
M.~J. Sandel.
\newblock \emph{What money can’t buy: the moral limits of markets}.
\newblock Brasenose College, Oxford, 1998.

\bibitem[Sanz(2016)]{sanz2016community}
E.~O. Sanz.
\newblock Community currency (ccs) in spain: An empirical study of their social effects.
\newblock \emph{Ecological Economics}, 121:\penalty0 20--27, 2016.

\bibitem[Sastry et~al.(2024)Sastry, Heim, Belfield, Anderljung, Brundage, Hazell, O'Keefe, Hadfield, Ngo, Pilz, et~al.]{sastry2024computing}
G.~Sastry, L.~Heim, H.~Belfield, M.~Anderljung, M.~Brundage, J.~Hazell, C.~O'Keefe, G.~K. Hadfield, R.~Ngo, K.~Pilz, et~al.
\newblock Computing power and the governance of artificial intelligence.
\newblock \emph{arXiv preprint arXiv:2402.08797}, 2024.

\bibitem[Satz(2010)]{satz2010some}
D.~Satz.
\newblock \emph{Why some things should not be for sale: The moral limits of markets}.
\newblock Oxford University Press, 2010.

\bibitem[Savona and Ciarli(2019)]{savona2019structural}
M.~Savona and T.~Ciarli.
\newblock Structural changes and sustainability. a selected review of the empirical evidence.
\newblock \emph{Ecological economics}, 159:\penalty0 244--260, 2019.

\bibitem[Schelling(1973)]{schelling1973hockey}
T.~C. Schelling.
\newblock Hockey helmets, concealed weapons, and daylight saving: A study of binary choices with externalities.
\newblock \emph{Journal of Conflict resolution}, 17\penalty0 (3):\penalty0 381--428, 1973.

\bibitem[Schill et~al.(2019)Schill, Anderies, Lindahl, Folke, Polasky, C{\'a}rdenas, Cr{\'e}pin, Janssen, Norberg, and Schl{\"u}ter]{schill2019more}
C.~Schill, J.~M. Anderies, T.~Lindahl, C.~Folke, S.~Polasky, J.~C. C{\'a}rdenas, A.-S. Cr{\'e}pin, M.~A. Janssen, J.~Norberg, and M.~Schl{\"u}ter.
\newblock A more dynamic understanding of human behaviour for the anthropocene.
\newblock \emph{Nature Sustainability}, 2\penalty0 (12):\penalty0 1075--1082, 2019.

\bibitem[Schneier(2024)]{schneier2024reimagining}
B.~Schneier.
\newblock Reimagining democracy.
\newblock \emph{Common Knowledge}, 30\penalty0 (3):\penalty0 354--358, 2024.

\bibitem[Schuldt(2012)]{schuldt2012multiagent}
A.~Schuldt.
\newblock Multiagent coordination enabling autonomous logistics.
\newblock \emph{KI-K{\"u}nstliche Intelligenz}, 26:\penalty0 91--94, 2012.

\bibitem[Schumpeter(1942)]{schumpeter1942capitalism}
J.~A. Schumpeter.
\newblock \emph{Capitalism, socialism and democracy}.
\newblock Harper and Brothers, 1942.

\bibitem[Schwarcz(2009)]{schwarcz2009regulating}
S.~L. Schwarcz.
\newblock Regulating complexity in financial markets.
\newblock \emph{Wash. UL Rev.}, 87:\penalty0 211, 2009.

\bibitem[Sedlmeir et~al.(2021)Sedlmeir, Smethurst, Rieger, and Fridgen]{sedlmeir2021digital}
J.~Sedlmeir, R.~Smethurst, A.~Rieger, and G.~Fridgen.
\newblock Digital identities and verifiable credentials.
\newblock \emph{Business \& Information Systems Engineering}, 63\penalty0 (5):\penalty0 603--613, 2021.

\bibitem[Seyfang(2006)]{seyfang2006sustainable}
G.~Seyfang.
\newblock Sustainable consumption, the new economics and community currencies: Developing new institutions for environmental governance.
\newblock \emph{Regional Studies}, 40\penalty0 (7):\penalty0 781--791, 2006.

\bibitem[Seyfang and Longhurst(2013)]{seyfang2013growing}
G.~Seyfang and N.~Longhurst.
\newblock Growing green money? mapping community currencies for sustainable development.
\newblock \emph{Ecological Economics}, 86:\penalty0 65--77, 2013.

\bibitem[Shah et~al.(2025)Shah, Irpan, Turner, Wang, Conmy, Lindner, Brown-Cohen, Ho, Nanda, Popa, et~al.]{shah2025approach}
R.~Shah, A.~Irpan, A.~M. Turner, A.~Wang, A.~Conmy, D.~Lindner, J.~Brown-Cohen, L.~Ho, N.~Nanda, R.~A. Popa, et~al.
\newblock An approach to technical agi safety and security.
\newblock \emph{arXiv preprint arXiv:2504.01849}, 2025.

\bibitem[Sharma et~al.(2023)Sharma, Tong, Korbak, Duvenaud, Askell, Bowman, Cheng, Durmus, Hatfield-Dodds, Johnston, et~al.]{sharma2023towards}
M.~Sharma, M.~Tong, T.~Korbak, D.~Duvenaud, A.~Askell, S.~R. Bowman, N.~Cheng, E.~Durmus, Z.~Hatfield-Dodds, S.~R. Johnston, et~al.
\newblock Towards understanding sycophancy in language models.
\newblock \emph{arXiv preprint arXiv:2310.13548}, 2023.

\bibitem[Shavit et~al.(2023)Shavit, Agarwal, Brundage, Adler, O’Keefe, Campbell, Lee, Mishkin, Eloundou, Hickey, et~al.]{shavit2023practices}
Y.~Shavit, S.~Agarwal, M.~Brundage, S.~Adler, C.~O’Keefe, R.~Campbell, T.~Lee, P.~Mishkin, T.~Eloundou, A.~Hickey, et~al.
\newblock Practices for governing agentic ai systems.
\newblock \emph{Research Paper, OpenAI}, 2023.

\bibitem[Shayegani et~al.(2023)Shayegani, Mamun, Fu, Zaree, Dong, and Abu-Ghazaleh]{shayegani2023survey}
E.~Shayegani, M.~A.~A. Mamun, Y.~Fu, P.~Zaree, Y.~Dong, and N.~Abu-Ghazaleh.
\newblock Survey of vulnerabilities in large language models revealed by adversarial attacks.
\newblock \emph{arXiv preprint arXiv:2310.10844}, 2023.

\bibitem[Shekhtman and Waisbard(2019)]{shekhtman2019engravechain}
L.~Shekhtman and E.~Waisbard.
\newblock Engravechain: Tamper-proof distributed log system.
\newblock In \emph{Proceedings of the 2nd Workshop on Blockchain-enabled Networked Sensor}, pages 8--14, 2019.

\bibitem[Siddarth et~al.(2020)Siddarth, Ivliev, Siri, and Berman]{siddarth2020watches}
D.~Siddarth, S.~Ivliev, S.~Siri, and P.~Berman.
\newblock Who watches the watchmen? a review of subjective approaches for sybil-resistance in proof of personhood protocols.
\newblock \emph{Frontiers in Blockchain}, 3:\penalty0 590171, 2020.

\bibitem[Singh et~al.(2025)Singh, Lane, Yu, Lu, Ramos, Cui, and Zhao]{singh2025generalized}
N.~Singh, S.~Lane, T.~Yu, J.~Lu, A.~Ramos, H.~Cui, and H.~Zhao.
\newblock A generalized platform for artificial intelligence-powered autonomous protein engineering.
\newblock \emph{bioRxiv}, pages 2025--02, 2025.

\bibitem[Siqueira et~al.(2020)Siqueira, Honig, Mariano, and Moraes]{siqueira2020commons}
A.~C.~O. Siqueira, B.~Honig, S.~Mariano, and J.~Moraes.
\newblock A commons strategy for promoting entrepreneurship and social capital: Implications for community currencies, cryptocurrencies, and value exchange.
\newblock \emph{Journal of Business Ethics}, 166\penalty0 (4):\penalty0 711--726, 2020.

\bibitem[Skalse et~al.(2022)Skalse, Howe, Krasheninnikov, and Krueger]{skalse2022defining}
J.~Skalse, N.~Howe, D.~Krasheninnikov, and D.~Krueger.
\newblock Defining and characterizing reward gaming.
\newblock \emph{Advances in Neural Information Processing Systems}, 35:\penalty0 9460--9471, 2022.

\bibitem[Smith et~al.(2025)Smith, Abdulhai, Diaz, Tesic, Trivedi, Vezhnevets, Hammond, Clifton, Chang, Duéñez-Guzmán, Agapiou, Matyas, and Leibo]{smith2025evaluating}
C.~Smith, M.~Abdulhai, M.~Diaz, M.~Tesic, R.~Trivedi, A.~S. Vezhnevets, L.~Hammond, J.~Clifton, M.~Chang, E.~A. Duéñez-Guzmán, J.~P. Agapiou, J.~Matyas, and J.~Z. Leibo.
\newblock Evaluating generalization capabilities of {LLM}-based agents in mixed-motive scenarios using concordia.
\newblock \emph{Advances in neural information processing systems}, 38, 2025.

\bibitem[Snyder and Swann~Jr(1978)]{snyder1978behavioral}
M.~Snyder and W.~B. Swann~Jr.
\newblock Behavioral confirmation in social interaction: From social perception to social reality.
\newblock \emph{Journal of experimental social psychology}, 14\penalty0 (2):\penalty0 148--162, 1978.

\bibitem[S{\"o}derholm(2020)]{soderholm2020green}
P.~S{\"o}derholm.
\newblock The green economy transition: the challenges of technological change for sustainability.
\newblock \emph{Sustainable Earth}, 3\penalty0 (1):\penalty0 6, 2020.

\bibitem[Stavins(2010)]{stavins2010market}
R.~N. Stavins.
\newblock Market-based environmental policies.
\newblock In \emph{Public policies for environmental protection}, pages 31--76. Routledge, 2010.

\bibitem[Stevens et~al.(2015)Stevens, Moray, and Bruneel]{stevens2015social}
R.~Stevens, N.~Moray, and J.~Bruneel.
\newblock The social and economic mission of social enterprises: Dimensions, measurement, validation, and relation.
\newblock \emph{Entrepreneurship theory and practice}, 39\penalty0 (5):\penalty0 1051--1082, 2015.

\bibitem[Stooke et~al.(2021)Stooke, Mahajan, Barros, Deck, Bauer, Sygnowski, Trebacz, Jaderberg, Mathieu, et~al.]{team2021open}
A.~Stooke, A.~Mahajan, C.~Barros, C.~Deck, J.~Bauer, J.~Sygnowski, M.~Trebacz, M.~Jaderberg, M.~Mathieu, et~al.
\newblock Open-ended learning leads to generally capable agents.
\newblock \emph{arXiv preprint arXiv:2107.12808}, 2021.

\bibitem[Strouse et~al.(2021)Strouse, McKee, Botvinick, Hughes, and Everett]{strouse2021collaborating}
D.~Strouse, K.~McKee, M.~Botvinick, E.~Hughes, and R.~Everett.
\newblock Collaborating with humans without human data.
\newblock \emph{Advances in neural information processing systems}, 34:\penalty0 14502--14515, 2021.

\bibitem[Strubell et~al.(2020)Strubell, Ganesh, and McCallum]{strubell2020energy}
E.~Strubell, A.~Ganesh, and A.~McCallum.
\newblock Energy and policy considerations for modern deep learning research.
\newblock In \emph{Proceedings of the AAAI conference on artificial intelligence}, volume~34, pages 13693--13696, 2020.

\bibitem[Tarashev et~al.(2009)Tarashev, Borio, and Tsatsaronis]{tarashev2009systemic}
N.~A. Tarashev, C.~E. Borio, and K.~Tsatsaronis.
\newblock The systemic importance of financial institutions.
\newblock \emph{BIS Quarterly Review, September}, 2009.

\bibitem[Tatom(2011)]{tatom2011financial}
J.~A. Tatom.
\newblock \emph{Financial Market Regulation Legislation and Implications}.
\newblock Springer, 2011.

\bibitem[team(2025)]{worldcoin}
W.~team.
\newblock Worldcoin, 2025.
\newblock URL \url{https://world.org/}.

\bibitem[Thomson(2011)]{thomson2011fair}
W.~Thomson.
\newblock Fair allocation rules.
\newblock In \emph{Handbook of social choice and welfare}, volume~2, pages 393--506. Elsevier, 2011.

\bibitem[Trevelyan et~al.(2016)Trevelyan, Hamel, and Kang]{trevelyan2016robotics}
J.~Trevelyan, W.~R. Hamel, and S.-C. Kang.
\newblock Robotics in hazardous applications.
\newblock In \emph{Springer handbook of robotics}, pages 1521--1548. Springer, 2016.

\bibitem[Tu et~al.(2025)Tu, Schaekermann, Palepu, Saab, Freyberg, Tanno, Wang, Li, Amin, Cheng, et~al.]{tu2025towards}
T.~Tu, M.~Schaekermann, A.~Palepu, K.~Saab, J.~Freyberg, R.~Tanno, A.~Wang, B.~Li, M.~Amin, Y.~Cheng, et~al.
\newblock Towards conversational diagnostic artificial intelligence.
\newblock \emph{Nature}, pages 1--9, 2025.

\bibitem[Tukker and Ekins(2019)]{tukker2019concepts}
A.~Tukker and P.~Ekins.
\newblock Concepts fostering resource efficiency: a trade-off between ambitions and viability.
\newblock \emph{Ecological Economics}, 155:\penalty0 36--45, 2019.

\bibitem[Vashishth et~al.(2024)Vashishth, Sharma, Sharma, Kumar, Chaudhary, and Panwar]{vashishth2024intelligent}
T.~K. Vashishth, V.~Sharma, K.~K. Sharma, B.~Kumar, S.~Chaudhary, and R.~Panwar.
\newblock Intelligent resource allocation and optimization for industrial robotics using {AI} and blockchain.
\newblock In \emph{AI and blockchain applications in industrial robotics}, pages 82--110. IGI Global Scientific Publishing, 2024.

\bibitem[Vedadi et~al.(2025)Vedadi, Barrett, Harris, Wulczyn, Reddy, Ruparel, Schaekermann, Strother, Tanno, Sharma, Lee, Hughes, Slack, Palepu, Freyberg, Saab, Liévin, Weng, Tu, Liu, Tomasev, Kulkarni, Mahdavi, Guu, Barral, Webster, Manyika, Hassidim, Chou, Matias, Kohli, Rodman, Natarajan, Karthikesalingam, and Stutz]{vedadi2025physiciancenteredoversightconversationaldiagnostic}
E.~Vedadi, D.~Barrett, N.~Harris, E.~Wulczyn, S.~Reddy, R.~Ruparel, M.~Schaekermann, T.~Strother, R.~Tanno, Y.~Sharma, J.~Lee, C.~Hughes, D.~Slack, A.~Palepu, J.~Freyberg, K.~Saab, V.~Liévin, W.-H. Weng, T.~Tu, Y.~Liu, N.~Tomasev, K.~Kulkarni, S.~S. Mahdavi, K.~Guu, J.~Barral, D.~R. Webster, J.~Manyika, A.~Hassidim, K.~Chou, Y.~Matias, P.~Kohli, A.~Rodman, V.~Natarajan, A.~Karthikesalingam, and D.~Stutz.
\newblock Towards physician-centered oversight of conversational diagnostic ai, 2025.
\newblock URL \url{https://arxiv.org/abs/2507.15743}.

\bibitem[Vinyals et~al.(2019)Vinyals, Babuschkin, Czarnecki, Mathieu, Dudzik, Chung, Choi, Powell, Ewalds, Georgiev, et~al.]{vinyals2019grandmaster}
O.~Vinyals, I.~Babuschkin, W.~M. Czarnecki, M.~Mathieu, A.~Dudzik, J.~Chung, D.~H. Choi, R.~Powell, T.~Ewalds, P.~Georgiev, et~al.
\newblock Grandmaster level in starcraft ii using multi-agent reinforcement learning.
\newblock \emph{nature}, 575\penalty0 (7782):\penalty0 350--354, 2019.

\bibitem[Vuorenmaa and Wang(2014)]{vuorenmaa2014agent}
T.~A. Vuorenmaa and L.~Wang.
\newblock An agent-based model of the flash crash of may 6, 2010, with policy implications.
\newblock \emph{Available at SSRN 2336772}, 2014.

\bibitem[Walzer(2008)]{walzer2008spheres}
M.~Walzer.
\newblock \emph{Spheres of justice: A defense of pluralism and equality}.
\newblock Basic books, 2008.

\bibitem[Wang et~al.(2025)Wang, Zhao, Jiang, Chen, Zhu, Chen, Liu, Zhang, Fan, Ma, et~al.]{wang2025beyond}
C.~Wang, Z.~Zhao, Y.~Jiang, Z.~Chen, C.~Zhu, Y.~Chen, J.~Liu, L.~Zhang, X.~Fan, H.~Ma, et~al.
\newblock Beyond reward hacking: Causal rewards for large language model alignment.
\newblock \emph{arXiv preprint arXiv:2501.09620}, 2025.

\bibitem[Wang et~al.(2014)Wang, Gao, Alsaadi, and Hayat]{wang2014overview}
Q.~Wang, H.~Gao, F.~Alsaadi, and T.~Hayat.
\newblock An overview of consensus problems in constrained multi-agent coordination.
\newblock \emph{Systems Science \& Control Engineering: An Open Access Journal}, 2\penalty0 (1):\penalty0 275--284, 2014.

\bibitem[Wang et~al.(2022)Wang, Gleave, Belrose, Tseng, Dennis, Duan, Pogrebniak, Miller, Levine, and Russell]{wang2022adversarial}
T.~T. Wang, A.~Gleave, N.~Belrose, T.~Tseng, M.~D. Dennis, Y.~Duan, V.~Pogrebniak, J.~Miller, S.~Levine, and S.~Russell.
\newblock Adversarial policies beat professional-level go ais.
\newblock In \emph{Deep Reinforcement Learning Workshop NeurIPS 2022}, 2022.

\bibitem[Wang et~al.(2023{\natexlab{a}})Wang, Gleave, Tseng, Pelrine, Belrose, Miller, Dennis, Duan, Pogrebniak, Levine, et~al.]{wang2023adversarial}
T.~T. Wang, A.~Gleave, T.~Tseng, K.~Pelrine, N.~Belrose, J.~Miller, M.~D. Dennis, Y.~Duan, V.~Pogrebniak, S.~Levine, et~al.
\newblock Adversarial policies beat superhuman go {AI}s.
\newblock In \emph{International Conference on Machine Learning}, pages 35655--35739. PMLR, 2023{\natexlab{a}}.

\bibitem[Wang et~al.(2023{\natexlab{b}})Wang, Cai, Chen, Liu, Ma, and Liang]{wang2023describe}
Z.~Wang, S.~Cai, G.~Chen, A.~Liu, X.~S. Ma, and Y.~Liang.
\newblock Describe, explain, plan and select: interactive planning with llms enables open-world multi-task agents.
\newblock \emph{Advances in Neural Information Processing Systems}, 36:\penalty0 34153--34189, 2023{\natexlab{b}}.

\bibitem[Wegner(1987)]{wegner1987transactive}
D.~M. Wegner.
\newblock Transactive memory: A contemporary analysis of the group mind.
\newblock In \emph{Theories of group behavior}, pages 185--208. Springer, 1987.

\bibitem[Wegner and Ward(2013)]{wegner2013internet}
D.~M. Wegner and A.~F. Ward.
\newblock The internet has become the external hard drive for our memories.
\newblock \emph{Scientific American}, 309\penalty0 (6):\penalty0 58--61, 2013.

\bibitem[Wijngaards et~al.(2002)Wijngaards, Overeinder, van Steen, and Brazier]{wijngaards2002supporting}
N.~J. Wijngaards, B.~J. Overeinder, M.~van Steen, and F.~M. Brazier.
\newblock Supporting internet-scale multi-agent systems.
\newblock \emph{Data \& Knowledge Engineering}, 41\penalty0 (2-3):\penalty0 229--245, 2002.

\bibitem[Wu et~al.(2024{\natexlab{a}})Wu, Koh, Salakhutdinov, Fried, and Raghunathan]{wu2024adversarial}
C.~H. Wu, J.~Y. Koh, R.~Salakhutdinov, D.~Fried, and A.~Raghunathan.
\newblock Adversarial attacks on multimodal agents.
\newblock \emph{arXiv e-prints}, pages arXiv--2406, 2024{\natexlab{a}}.

\bibitem[Wu et~al.(2022)Wu, Raghavendra, Gupta, Acun, Ardalani, Maeng, Chang, Aga, Huang, Bai, et~al.]{wu2022sustainable}
C.-J. Wu, R.~Raghavendra, U.~Gupta, B.~Acun, N.~Ardalani, K.~Maeng, G.~Chang, F.~Aga, J.~Huang, C.~Bai, et~al.
\newblock Sustainable ai: Environmental implications, challenges and opportunities.
\newblock \emph{Proceedings of Machine Learning and Systems}, 4:\penalty0 795--813, 2022.

\bibitem[Wu et~al.(2016)Wu, Balliet, and Van~Lange]{wu2016reputation}
J.~Wu, D.~Balliet, and P.~A. Van~Lange.
\newblock Reputation, gossip, and human cooperation.
\newblock \emph{Social and Personality Psychology Compass}, 10\penalty0 (6):\penalty0 350--364, 2016.

\bibitem[Wu et~al.(2025)Wu, Wang, Wang, Dong, Yang, Li, Huang, Zhao, Li, Wang, Fan, and Chen]{wu2025ainativeexperimentallaboratoryautonomous}
M.~Wu, Z.~Wang, J.~Wang, Z.~Dong, J.~Yang, Q.~Li, T.~Huang, L.~Zhao, M.~Li, F.~Wang, C.~Fan, and H.~Chen.
\newblock An ai-native experimental laboratory for autonomous biomolecular engineering, 2025.
\newblock URL \url{https://arxiv.org/abs/2507.02379}.

\bibitem[Wu et~al.(2024{\natexlab{b}})Wu, Sun, Li, Welleck, and Yang]{wu2024inference}
Y.~Wu, Z.~Sun, S.~Li, S.~Welleck, and Y.~Yang.
\newblock Inference scaling laws: An empirical analysis of compute-optimal inference for problem-solving with language models.
\newblock \emph{arXiv preprint arXiv:2408.00724}, 2024{\natexlab{b}}.

\bibitem[Xi et~al.(2025)Xi, Chen, Guo, He, Ding, Hong, Zhang, Wang, Jin, Zhou, et~al.]{xi2025rise}
Z.~Xi, W.~Chen, X.~Guo, W.~He, Y.~Ding, B.~Hong, M.~Zhang, J.~Wang, S.~Jin, E.~Zhou, et~al.
\newblock The rise and potential of large language model based agents: A survey.
\newblock \emph{Science China Information Sciences}, 68\penalty0 (2):\penalty0 121101, 2025.

\bibitem[Xu et~al.(2024{\natexlab{a}})Xu, Almahri, Mak, and Brintrup]{xu2024multi}
L.~Xu, S.~Almahri, S.~Mak, and A.~Brintrup.
\newblock Multi-agent systems and foundation models enable autonomous supply chains: Opportunities and challenges.
\newblock \emph{IFAC-PapersOnLine}, 58\penalty0 (19):\penalty0 795--800, 2024{\natexlab{a}}.

\bibitem[Xu et~al.(2024{\natexlab{b}})Xu, Mak, Minaricova, and Brintrup]{xu2024implementing}
L.~Xu, S.~Mak, M.~Minaricova, and A.~Brintrup.
\newblock On implementing autonomous supply chains: A multi-agent system approach.
\newblock \emph{Computers in Industry}, 161:\penalty0 104120, 2024{\natexlab{b}}.

\bibitem[Xue(2023)]{xue2023strategies}
D.~Xue.
\newblock Strategies for mitigating the global energy and carbon impact of artificial intelligence.
\newblock \emph{United Nations SDGs}, pages 2023--05, 2023.

\bibitem[Yan et~al.(2023)Yan, Guo, Lou, Wang, Zhang, and Du]{yan2023efficient}
X.~Yan, J.~Guo, X.~Lou, J.~Wang, H.~Zhang, and Y.~Du.
\newblock An efficient end-to-end training approach for zero-shot human-ai coordination.
\newblock \emph{Advances in Neural Information Processing Systems}, 36:\penalty0 2636--2658, 2023.

\bibitem[Yang and Zhai(2025)]{yang2025principlesaiagenteconomics}
K.~Yang and C.~Zhai.
\newblock Ten principles of {AI} agent economics, 2025.
\newblock URL \url{https://arxiv.org/abs/2505.20273}.

\bibitem[Yang et~al.(2024{\natexlab{a}})Yang, Pan, Luo, Qiu, Zhong, Yu, and Chen]{yang2024rewards}
R.~Yang, X.~Pan, F.~Luo, S.~Qiu, H.~Zhong, D.~Yu, and J.~Chen.
\newblock Rewards-in-context: Multi-objective alignment of foundation models with dynamic preference adjustment.
\newblock \emph{arXiv preprint arXiv:2402.10207}, 2024{\natexlab{a}}.

\bibitem[Yang et~al.(2025)Yang, Wen, Wang, and Zhang]{yang2025agentexchangeshapingfuture}
Y.~Yang, Y.~Wen, J.~Wang, and W.~Zhang.
\newblock Agent exchange: Shaping the future of {AI} agent economics, 2025.
\newblock URL \url{https://arxiv.org/abs/2507.03904}.

\bibitem[Yang et~al.(2024{\natexlab{b}})Yang, Liu, Liu, Liu, Xiong, Wang, Yang, Hu, Chen, Zhang, et~al.]{yang2024position}
Z.~Yang, A.~Liu, Z.~Liu, K.~Liu, F.~Xiong, Y.~Wang, Z.~Yang, Q.~Hu, X.~Chen, Z.~Zhang, et~al.
\newblock Position: Towards unified alignment between agents, humans, and environment.
\newblock In \emph{Forty-first International Conference on Machine Learning}, 2024{\natexlab{b}}.

\bibitem[Yao et~al.(2024)Yao, Duan, Xu, Cai, Sun, and Zhang]{yao2024survey}
Y.~Yao, J.~Duan, K.~Xu, Y.~Cai, Z.~Sun, and Y.~Zhang.
\newblock A survey on large language model (llm) security and privacy: The good, the bad, and the ugly.
\newblock \emph{High-Confidence Computing}, page 100211, 2024.

\bibitem[Zech(2021)]{zech2021liability}
H.~Zech.
\newblock Liability for {AI}: public policy considerations.
\newblock In \emph{ERA forum}, volume~22, pages 147--158. Springer, 2021.

\bibitem[Zhang et~al.(2022)Zhang, Lou, Sun, Su, and Li]{zhang2022truthful}
J.~Zhang, W.~Lou, H.~Sun, Q.~Su, and W.~Li.
\newblock Truthful auction mechanisms for resource allocation in the internet of vehicles with public blockchain networks.
\newblock \emph{Future Generation Computer Systems}, 132:\penalty0 11--24, 2022.

\bibitem[Zhang et~al.(2023)Zhang, Liu, Qin, Xu, and Zhang]{zhang2023adaptive}
J.~Zhang, Y.~Liu, X.~Qin, X.~Xu, and P.~Zhang.
\newblock Adaptive resource allocation for blockchain-based federated learning in internet of things.
\newblock \emph{IEEE Internet of Things Journal}, 10\penalty0 (12):\penalty0 10621--10635, 2023.

\bibitem[Zhang et~al.(2014)Zhang, Liang, Lu, and Shen]{zhang2014sybil}
K.~Zhang, X.~Liang, R.~Lu, and X.~Shen.
\newblock Sybil attacks and their defenses in the internet of things.
\newblock \emph{IEEE Internet of Things Journal}, 1\penalty0 (5):\penalty0 372--383, 2014.

\bibitem[Zhang et~al.(2021)Zhang, Yang, and Ba{\c{s}}ar]{zhang2021multi}
K.~Zhang, Z.~Yang, and T.~Ba{\c{s}}ar.
\newblock Multi-agent reinforcement learning: A selective overview of theories and algorithms.
\newblock \emph{Handbook of reinforcement learning and control}, pages 321--384, 2021.

\bibitem[Zhang et~al.(2024)Zhang, Bai, Wang, Ye, Ma, and Yang]{zhang2024incentive}
Z.~Zhang, F.~Bai, M.~Wang, H.~Ye, C.~Ma, and Y.~Yang.
\newblock Incentive compatibility for {AI} alignment in sociotechnical systems: Positions and prospects.
\newblock \emph{arXiv preprint arXiv:2402.12907}, 2024.

\bibitem[Zhao et~al.(2023)Zhao, Song, Yuan, Hu, Gao, Wu, Sun, and Yang]{zhao2023maximum}
R.~Zhao, J.~Song, Y.~Yuan, H.~Hu, Y.~Gao, Y.~Wu, Z.~Sun, and W.~Yang.
\newblock Maximum entropy population-based training for zero-shot human-ai coordination.
\newblock In \emph{Proceedings of the AAAI Conference on Artificial Intelligence}, volume~37, pages 6145--6153, 2023.

\bibitem[Zheng et~al.(2023)Zheng, Chiang, Sheng, Zhuang, Wu, Zhuang, Lin, Li, Li, Xing, et~al.]{zheng2023judging}
L.~Zheng, W.-L. Chiang, Y.~Sheng, S.~Zhuang, Z.~Wu, Y.~Zhuang, Z.~Lin, Z.~Li, D.~Li, E.~Xing, et~al.
\newblock Judging llm-as-a-judge with mt-bench and chatbot arena.
\newblock \emph{Advances in Neural Information Processing Systems}, 36:\penalty0 46595--46623, 2023.

\bibitem[Zheng et~al.(2022)Zheng, Trott, Srinivasa, Parkes, and Socher]{zheng2022ai}
S.~Zheng, A.~Trott, S.~Srinivasa, D.~C. Parkes, and R.~Socher.
\newblock The {AI} economist: Taxation policy design via two-level deep multiagent reinforcement learning.
\newblock \emph{Science advances}, 8\penalty0 (18):\penalty0 eabk2607, 2022.

\bibitem[Zheng et~al.(2018)Zheng, Xie, Dai, Chen, and Wang]{zheng2018blockchain}
Z.~Zheng, S.~Xie, H.-N. Dai, X.~Chen, and H.~Wang.
\newblock Blockchain challenges and opportunities: A survey.
\newblock \emph{International journal of web and grid services}, 14\penalty0 (4):\penalty0 352--375, 2018.

\bibitem[Zhou et~al.(2024{\natexlab{a}})Zhou, Diro, Saini, Kaisar, and Hiep]{zhou2024leveraging}
L.~Zhou, A.~Diro, A.~Saini, S.~Kaisar, and P.~C. Hiep.
\newblock Leveraging zero knowledge proofs for blockchain-based identity sharing: A survey of advancements, challenges and opportunities.
\newblock \emph{Journal of Information Security and Applications}, 80:\penalty0 103678, 2024{\natexlab{a}}.

\bibitem[Zhou et~al.(2024{\natexlab{b}})Zhou, Song, Yao, Shu, and Ma]{zhou2024isr}
Z.~Zhou, J.~Song, K.~Yao, Z.~Shu, and L.~Ma.
\newblock Isr-llm: Iterative self-refined large language model for long-horizon sequential task planning.
\newblock In \emph{2024 IEEE International Conference on Robotics and Automation (ICRA)}, pages 2081--2088. IEEE, 2024{\natexlab{b}}.

\bibitem[Zhu et~al.(2023{\natexlab{a}})Zhu, Wang, Zhou, Wang, Chen, Wang, Yang, Ye, Zhang, Gong, et~al.]{zhu2023promptrobust}
K.~Zhu, J.~Wang, J.~Zhou, Z.~Wang, H.~Chen, Y.~Wang, L.~Yang, W.~Ye, Y.~Zhang, N.~Gong, et~al.
\newblock Promptrobust: Towards evaluating the robustness of large language models on adversarial prompts.
\newblock In \emph{Proceedings of the 1st ACM Workshop on Large {AI} Systems and Models with Privacy and Safety Analysis}, pages 57--68, 2023{\natexlab{a}}.

\bibitem[Zhu et~al.(2023{\natexlab{b}})Zhu, Wang, Zhou, Wang, Chen, Wang, Yang, Ye, Zhang, Zhenqiang~Gong, et~al.]{zhu2023promptbench}
K.~Zhu, J.~Wang, J.~Zhou, Z.~Wang, H.~Chen, Y.~Wang, L.~Yang, W.~Ye, Y.~Zhang, N.~Zhenqiang~Gong, et~al.
\newblock Promptbench: Towards evaluating the robustness of large language models on adversarial prompts.
\newblock \emph{arXiv e-prints}, pages arXiv--2306, 2023{\natexlab{b}}.

\bibitem[Zhu et~al.(2021)Zhu, Hu, Li, and Zhu]{zhu2021using}
P.~Zhu, J.~Hu, X.~Li, and Q.~Zhu.
\newblock Using blockchain technology to enhance the traceability of original achievements.
\newblock \emph{IEEE Transactions on Engineering Management}, 70\penalty0 (5):\penalty0 1693--1707, 2021.

\bibitem[Zhu et~al.(2025)Zhu, Sun, Nian, South, Pentland, and Pei]{zhu2025automatedriskygamemodeling}
S.~Zhu, J.~Sun, Y.~Nian, T.~South, A.~Pentland, and J.~Pei.
\newblock The automated but risky game: Modeling agent-to-agent negotiations and transactions in consumer markets, 2025.
\newblock URL \url{https://arxiv.org/abs/2506.00073}.

\bibitem[Zhuang and Hadfield-Menell(2020)]{zhuang2020consequences}
S.~Zhuang and D.~Hadfield-Menell.
\newblock Consequences of misaligned ai.
\newblock \emph{Advances in Neural Information Processing Systems}, 33:\penalty0 15763--15773, 2020.

\bibitem[Zhuge et~al.(2025)Zhuge, Zhao, Ashley, Wang, Khizbullin, Xiong, Liu, Chang, Krishnamoorthi, Tian, et~al.]{zhuge2025agent}
M.~Zhuge, C.~Zhao, D.~R. Ashley, W.~Wang, D.~Khizbullin, Y.~Xiong, Z.~Liu, E.~Chang, R.~Krishnamoorthi, Y.~Tian, et~al.
\newblock Agent-as-a-judge: Evaluating agents with agents.
\newblock 2025.

\bibitem[Zimmer et~al.(2021)Zimmer, Glanois, Siddique, and Weng]{zimmer2021learning}
M.~Zimmer, C.~Glanois, U.~Siddique, and P.~Weng.
\newblock Learning fair policies in decentralized cooperative multi-agent reinforcement learning.
\newblock In \emph{International conference on machine learning}, pages 12967--12978. PMLR, 2021.

\bibitem[Zou et~al.(2019)Zou, Lo, Kochhar, Le, Xia, Feng, Chen, and Xu]{zou2019smart}
W.~Zou, D.~Lo, P.~S. Kochhar, X.-B.~D. Le, X.~Xia, Y.~Feng, Z.~Chen, and B.~Xu.
\newblock Smart contract development: Challenges and opportunities.
\newblock \emph{IEEE transactions on software engineering}, 47\penalty0 (10):\penalty0 2084--2106, 2019.

\end{thebibliography}

\section*{Acknowledgements}
We would like to thank our colleagues who provided feedback on the manuscript, in particular Sébastien Krier and Raphael Koster for their valuable comments.

\end{document}